# Max-Sliced Wasserstein Distance and its use for GANs

Ishan Deshpande, Yuan-Ting Hu, Ruoyu Sun, Ayis Pyrros[†], Nasir Siddiqui[†],
Sanmi Koyejo, Zhizhen Zhao, David Forsyth, Alexander Schwing
University of Illinois at Urbana-Champaign [†]Dupage Medical Group
ishan.sd@gmail.com, {ythu2, ruoyus}@illinois.edu, ayis@ayis.org, nsiddiqui@gmail.com,
{sanmi, zhizhenz, daf, aschwing}@illinois.edu

## Abstract

*Generative adversarial nets (GANs) and variational auto-encoders have significantly improved our distribution modeling capabilities, showing promise for dataset augmentation, image-to-image translation and feature learning. However, to model high-dimensional distributions, sequential training and stacked architectures are common, increasing the number of tunable hyper-parameters as well as the training time. Nonetheless, the sample complexity of the distance metrics remains one of the factors affecting GAN training. We first show that the recently proposed sliced Wasserstein distance has compelling sample complexity properties when compared to the Wasserstein distance. To further improve the sliced Wasserstein distance we then analyze its 'projection complexity' and develop the max-sliced Wasserstein distance which enjoys compelling sample complexity while reducing projection complexity, albeit necessitating a max estimation. We finally illustrate that the proposed distance trains GANs on high-dimensional images up to a resolution of 256x256 easily.*

## 1. Introduction

Generative modeling capabilities have improved tremendously in the last few years, especially since the advent of deep learning-based models like generative adversarial nets (GANs) [11] and variational auto-encoders (VAEs) [17]. Instead of sampling from a high-dimensional distribution, GANs and VAEs transform a sample obtained from a simple distribution using deep nets. These models have found use in dataset augmentation [31], image-to-image translation [15, 37, 21, 14, 24, 29, 35, 38], and even feature learning for inference related tasks [9].

GANs and many of their variants formulate generative modeling as a two player game. A 'generator' creates samples that resemble the ground truth data. A 'discriminator' tries to distinguish between 'artificial' and 'real' samples. Both, the generator and discriminator, are parametrized using deep nets and trained via stochastic gradient descent. In its original formulation [11], a GAN minimizes the Jenson-Shannon divergence between the data distribution and the probability distribution induced in the data space by the generator. Many other variants have been proposed, which use either some divergence or the integral probability metric to measure the distance between the distributions [2, 22, 12, 20, 8, 7, 27, 4, 26, 23, 13, 30]. When carefully trained, GANs are able to produce high quality samples [28, 16, 25, 16, 25]. Training GANs is, however, difficult – especially on high dimensional datasets.

The scaling difficulty of GANs may be related to one fundamental theoretical issue: the sample complexity. It is shown in [3] that KL-divergence, Jenson-Shannon and Wasserstein distance do not generalize, in the sense that the population distance cannot be approximated by an empirical distance when there are only a polynomial number of samples. To improve generalization, one popular method is to limit the discriminator class [3, 10] and interpret the training process as minimizing a neural-net distance [3].

In this work, we promote a different path that resolves the sample complexity issue. A fundamental reason for the exponential sample complexity of the Wasserstein distance is the sparsity of points in a high dimensional space. Even if two collections of points are randomly drawn from the same ball, these two collections are far away from each other. Our intuition is that projection onto a low-dimensional subspace, such as a line, mitigates the artificial distance effect in high dimensions and the distance of the projected samples reflects the true distance.

We first apply this intuition to analyze the recently proposed sliced Wasserstein distance GAN, which is based on the average Wasserstein distance of the projected versions of two distributions along a few randomly picked directions [8, 20, 34]. We prove that the sliced Wasserstein distance is generalizable for Gaussian distributions (*i.e.*, it has polynomial sample complexity), while Wasserstein distance is not, thus partially explaining why [8, 20, 34] may exhibit better behavior than the Wasserstein distance [2].

One drawback of the sliced Wasserstein distance is that it requires a large number of projection directions, since random directions lose a lot of information. To address this concern, we propose to project onto the "best direction," along which the projected distance is maximized. We call the corresponding metric the "max-sliced Wasserstein distance," and prove that it is also generalizable for Gaussian distributions.

Using this new metric, we are able to train GANs to generate high resolution images from the CelebA-HQ [16] and LSUN Bedrooms [36] datasets. We also achieve improved performance in other distribution matching tasks like unpaired word translation [6].

The main contributions of this paper are the following:

- We analyze in Sec. 3.1 the sample complexity of the Wasserstein and sliced Wasserstein distances. We show that for a certain class of distributions the Wasserstein distance has an exponential sample complexity, while the sliced Wasserstein distance [8, 34] has a polynomial sample complexity.
- We then study in Sec. 3.2 the projection complexity of the sliced Wasserstein distance, *i.e.*, how the number of random projection directions affects estimation.
- We introduce the max-sliced Wasserstein distance in Sec. 3.3 to address the projection complexity issue.
- We then employ the max-sliced Wasserstein distance to train GANs in Sec. 4, demonstrating significant reduction in the number of projection directions required for the sliced-Wasserstein GAN.

## 2. Background

Generative modeling is the task of learning a probability distribution from a given dataset $\mathcal{D} = \{(x)\}$ of samples $x \sim \mathbb{P}_d$ drawn from an unknown data distribution $\mathbb{P}_d$. While this has traditionally been seen through the lens of likelihood-maximization, GANs pose generative modeling as a distance minimization problem. More specifically, these approaches recommend learning the data distribution $\mathbb{P}_d$ by finding a distribution $\mathbb{P}_g$ that solves:

$$\operatorname*{argmin}_{\mathbb{P}_g} D(\mathbb{P}_g, \mathbb{P}_d), \tag{1}$$

where $D(\cdot, \cdot)$ is some distance or divergence between distributions. Arjovsky *et al.* [1] proposed using the Wasserstein distance in the context of GAN formulations. The Wasserstein-p distance between distributions $\mathbb{P}_g$ and $\mathbb{P}_d$ is defined as:

$$W_p(\mathbb{P}_g, \mathbb{P}_d) = \inf_{\gamma \in \Pi(\mathbb{P}_g, \mathbb{P}_d)} (\mathbb{E}_{(x,y)\sim\gamma}[||x - y||^p])^{\frac{1}{p}}, \tag{2}$$

where $\Pi(\mathbb{P}_g, \mathbb{P}_d)$ is the set of all possible joint distributions on $(x, y)$ with marginals $\mathbb{P}_g$ and $\mathbb{P}_d$.

Estimating the Wasserstein distance is, however, not straightforward. Arjovsky *et al.* [2] used the Kantorovich-Rubinstein duality to the Wasserstein-1 distance, which states that:

$$W(\mathbb{P}_g, \mathbb{P}_d) = \sup_{\|f\|_L \le 1} \mathbb{E}_{x\sim\mathbb{P}_g}[f(x)] - \mathbb{E}_{x\sim\mathbb{P}_d}[f(x)], \tag{3}$$

where the supremum is over all 1-Lipschitz functions $f : \mathcal{X} \to \mathbb{R}$. The function $f$ is commonly represented via a deep net and various ways have been suggested to enforce the Lipschitz constraint, *e.g.*, [12].

While the Wasserstein distance based approaches have been successful in several complex generative tasks, they suffer from instability arising from incorrect estimation. The cause behind this was noted in [33], where it was shown that estimates of the Wasserstein distance suffer from the 'curse of dimensionality.' To tackle the instability and complexity, a sliced version of the Wasserstein-2 distance was employed by [8, 20, 18, 34], which only requires estimating distances of 1-d distributions and is, therefore, more efficient. The "sliced Wasserstein-p distance" [5] between distributions $\mathbb{P}_d$ and $\mathbb{P}_g$ is defined as

$$\tilde{W}_p(\mathbb{P}_d, \mathbb{P}_g) = \left[\int_{\omega\in\Omega} W_p^p(\mathbb{P}_d^\omega, \mathbb{P}_g^\omega) d\omega\right]^{\frac{1}{p}}, \tag{4}$$

where $\mathbb{P}_g^\omega$, $\mathbb{P}_d^\omega$ denote the projection (*i.e.*, marginal) of $\mathbb{P}_g$, $\mathbb{P}_d$ onto the direction $\omega$, and $\Omega$ is the set of all possible directions on the unit sphere. Kolouri *et al.* [19] have shown that the sliced Wasserstein distance satisfies the properties of non-negativity, identity of indiscernibles, symmetry, and subadditivity. Hence, it is a true metric.

In practice, Deshpande *et al.* [8] approximate the sliced Wasserstein-2 distance between the distributions by using samples $\mathcal{D} \sim \mathbb{P}_d$, $\mathcal{F} \sim \mathbb{P}_g$, and a finite number of random Gaussian directions, replacing the integration over $\Omega$ with a summation over a randomly chosen set of unit vectors $\hat{\Omega} \propto \mathcal{N}(0, I)$, where '$\propto$' is used to indicate normalization to unit length. With $\mathbb{P}_g$ (and hence, $\mathcal{F}$) being implicitly parametrized by $\theta_g$, [8] uses the following program for generative modeling:

$$\min_{\theta_g} \frac{1}{|\hat{\Omega}|} \sum_{\omega\in\hat{\Omega}} W_2^2(\mathcal{D}^\omega, \mathcal{F}^\omega). \tag{5}$$

The Wasserstein-2 distance between the projected samples $\mathcal{D}^\omega$ and $\mathcal{F}^\omega$ can be computed by finding the optimal transport map. For 1-d distributions, this can be done through sorting [32], *i.e.*,

$$W_2^2(\mathcal{D}^\omega, \mathcal{F}^\omega) = \frac{1}{|\mathcal{D}|} \sum_i ||\mathcal{D}^\omega_{\pi_\mathcal{D}(i)} - \mathcal{F}^\omega_{\pi_\mathcal{F}(i)}||_2^2, \tag{6}$$

where $\pi_{\mathcal{D}}$ and $\pi_{\mathcal{F}}$ are permutations that sort the projected sample sets $\mathcal{D}^\omega$ and $\mathcal{F}^\omega$ respectively, *i.e.*, $\mathcal{D}^\omega_{\pi_{\mathcal{D}}(1)} \leq \mathcal{D}^\omega_{\pi_{\mathcal{D}}(2)} \leq \cdots \leq \mathcal{D}^\omega_{\pi_{\mathcal{D}}(|\mathcal{D}|)}$.

The program in Eq. (5), when coupled with a discriminator, was shown to work well on high-dimensional datasets. Instead of working directly with sets $\mathcal{D}$ and $\mathcal{F}$, it was proposed that we transform them to an adversarially learnt feature space, say $h_{\mathcal{D}}$ and $h_{\mathcal{F}}$ respectively, where $h$ is implicitly parameterized by $\theta_d$, *e.g.*, by using a deep net. The generator, parametrized by $\theta_g$, minimizes

$$\min_{\theta_g} \frac{1}{|\hat{\Omega}|} \sum_{\omega \in \hat{\Omega}} W_2^2(h^\omega_{\mathcal{D}}, h^\omega_{\mathcal{F}}). \tag{7}$$

The adversarial feature space $h$ is learnt via a discriminator which classifies real and fake data. This discriminator can be written as $\omega_d^T h$, where $\omega_d$ is a logistic layer and the parameters are learnt using

$$\hat{\theta}_d, \hat{\omega}_d = \operatorname*{argmax}_{\theta_d, \omega_d} \sum_{x \in \mathcal{D}} \ln(\sigma(\omega_d^T h_x)) + \sum_{\hat{x} \in \mathcal{F}} \ln(1 - \sigma(\omega_d^T h_{\hat{x}})). \tag{8}$$

# 3. Analysis and Max-Sliced Distance

In this section we provide the first analysis of the sample-complexity benefits of the sliced Wasserstein distance compared to the Wasserstein distance. We discuss how 'projection complexity' is a shortcoming of the sliced Wasserstein distance and present as a fix the max-sliced Wasserstein distance, which – as we will show – enjoys the same beneficial sample-complexity as the slice Wasserstein distance, albeit necessitating estimation of a maximum. We will then show how those results are used for training GANs.

## 3.1. Sample complexity of the Wasserstein and sliced Wasserstein distances

We first show the benefits of using the sliced Wasserstein distance over the Wasserstein distance. Specifically, we show that, in certain cases, estimation of the sliced Wasserstein distance has polynomial complexity, while the Wasserstein distance does not. To make this notion concrete, we introduce 'generalizability' of a distance:

**Definition 1** *Consider a family of distributions $\mathcal{P}$ over $\mathbb{R}^d$. A distance dist$(\cdot,\cdot)$ is said to be $\mathcal{P}$-generalizable if there exists a polynomial $g$ such that for any two distributions $\mu, \nu \in \mathcal{P}$, and their empirical ensembles $\hat{\mu}, \hat{\nu}$ with size $n = g(d, 1/\epsilon), \epsilon > 0$, the following holds:*

$$|dist(\mu,\nu) - dist(\hat{\mu},\hat{\nu})| \leq \epsilon \text{ w.p.} \geq 1 - polynomial(-n).$$

With this definition, we can prove the following result:

**Claim 1** *Consider the family of Gaussian distributions*

$$\mathcal{P} = \{\mathcal{N}(a, I) \mid a \in \mathbb{R}^d\}.$$

*The sliced Wasserstein-2 distance $\tilde{W}_2$ defined in Eq. (4) is $\mathcal{P}$-generalizable whereas the Wasserstein-2 distance $W_2$ defined in Eq. (2) is not.*

**Proof.** See the supplementary material. ■

Claim 1 implies that for GAN training, under certain conditions, it is better to use the sliced Wasserstein distance as we can get a more accurate training signal with a fixed computational budget. This will result in a more stable discriminator.

Even though the sliced Wasserstein distance enjoys better sample complexity, it has limitations when a finite number of random projection directions is used. We refer to this property as 'projection complexity' and illustrate it in the following section. We then present our proposed method to help alleviate this problem.

## 3.2. Projection complexity of the Sliced Wasserstein Distance

We begin with a simple example to demonstrate the limitations of using $\tilde{W}_2$ defined in Eq. (4) for learning distributions through gradient descent. To analyze the 'projection complexity' of $\tilde{W}_2$ we use infinitely many samples, but we use only finitely many directions $\omega \in \hat{\Omega}$.

Concretely, consider two $d$-dimensional Gaussians $\mu, \nu$ with identity covariance. Let $\mu = \mathcal{N}(0, I) = \mathbb{P}_d$ be the data distribution and let $\nu = \mathcal{N}(\beta\hat{e}, I) = \mathbb{P}_g$ be the induced generator distribution, parametrized only by its mean $\beta$, while $\hat{e}$ is a fixed unit vector. Using gradient descent on the estimated sliced Wasserstein distance between $\mu$ and $\nu$, we aim to learn $\beta$ so that $\mu = \nu$. Thus, the updates for $\beta$ are

$$\beta \leftarrow \beta - \alpha \nabla_\beta \tilde{W}_2(\mu, \nu), \tag{9}$$

where $\alpha$ is the learning rate.

The sliced Wasserstein distance $\tilde{W}_2$ is calculated by projecting the *distributions* (since we use infinitely many samples) onto random directions and comparing the projections, *i.e.*, marginals. Therefore, the estimated distance is

$$\tilde{W}_2(\mu, \nu) = \frac{1}{|\hat{\Omega}|} \sum_{\omega \in \hat{\Omega}} W_2(\mu^\omega, \nu^\omega), \tag{10}$$

where $W_2(\mu^\omega, \nu^\omega)$ is the Wasserstein distance between marginal distributions $\mu^\omega$, $\nu^\omega$. Note that each $\omega$ is normalized to unit norm.

Intuitively, projection of the Gaussians $\mu$, $\nu$ onto any direction other than $\hat{e}$ makes them appear closer than they actually are – making the learning process slower. For any

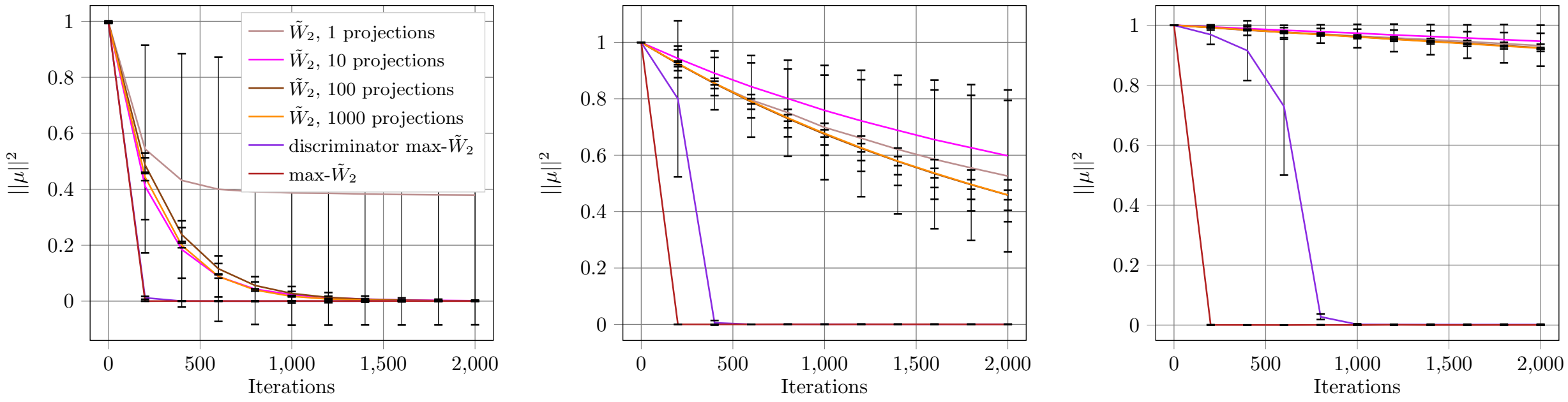

(a) $d = 10$ (b) $d = 100$ (c) $d = 1000$

Figure 1: Convergence of the mean for different sampling strategies for learning the mean of a $d$-dimensional Gaussian using the sliced Wasserstein distance and the max-sliced Wasserstein distance. Numbers in the legend denote the number of projection directions used.

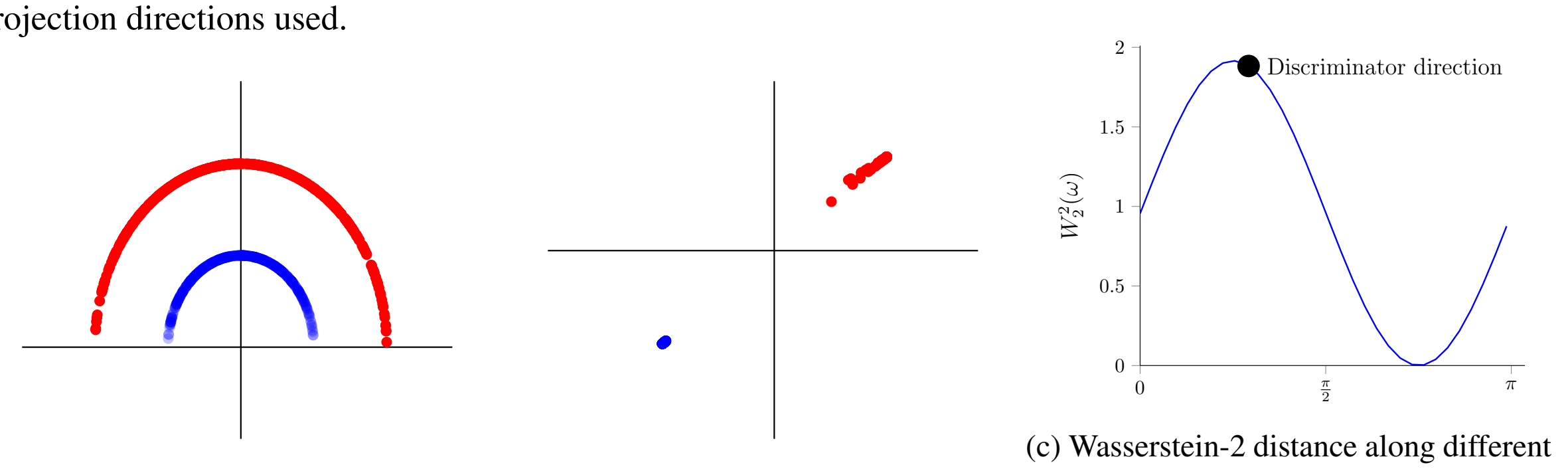

(a) Original distributions. (b) In feature space. (c) Wasserstein-2 distance along different projection angles (in radians) in the feature space.

Figure 2: The discriminator is able to identify important projection directions. The discriminator transforms the distributions in Fig. 2a to Fig. 2b. In this new space, the discriminator's direction is aligned with the one along which the distributions are the most dissimilar as shown in Fig. 2c.

given $\omega$, it is easy to see that $W_2(\mu^\omega, \nu^\omega) = \beta|\hat{e}^T\omega|$. Therefore, the update equation for $\beta$ is

$$\beta \to \beta - \alpha \frac{1}{|\hat{\Omega}|} \sum_{\omega \in \Omega} |\hat{e}^T \omega|. \tag{11}$$

The updates to $\beta$ are particularly small for high dimensional distributions, since any random unit-norm direction $\omega$ is orthogonal to $\hat{e}$ with high probability. Therefore, $\beta \to 0$ very slowly. We verify this effect empirically in Fig. 1, experimenting with different numbers of random projections and find that using the sliced Wasserstein distance results in very slow convergence. This problem is further aggravated when the dimensions of the distributions increase.

It is intuitively obvious that the aforementioned problem can easily be solved by choosing $\hat{e}$ as the projection direction. This results in larger updates and, consequently, faster convergence. This intuition is also verified empirically. We repeat the same experiment of learning $\beta$, but this time we use only one projection direction $\omega = \hat{e}$. This is labelled as max-$\tilde{W}_2$ in Fig. 1. By simply using the important projection direction, we achieve fast convergence of the mean.

Considering this example, it is evident that some projection directions are more meaningful than others. Therefore, GAN training should benefit from including such directions when comparing distributions. This observation motivates the max-sliced Wasserstein distance which we discuss next.

### 3.3. Max sliced Wasserstein distance

In this section we introduce the max-sliced Wasserstein distance and illustrate that it fixes the 'projection complexity' concern. We also prove that the max-sliced Wasserstein distance enjoys the same sample-complexity as the sliced Wasserstein distance, *i.e.*, we are not trading one benefit for another.

As noted in Sec. 3.2, it is useful to include the most meaningful projection direction. Formally, for the aforementioned example of $\mu = \mathcal{N}(0, I), \nu = \mathcal{N}(\beta\hat{e}, I)$, we want to use the direction $\omega^*$ that satisfies

$$\omega^* = \underset{\omega \in \Omega}{\operatorname{argmax}}\, |\hat{e}^T \omega|. \tag{12}$$

Comparing distributions along such a direction $\omega^*$ can, in fact, be shown to be a proper distance. We call it the 'max-sliced Wasserstein distance' and define it as follows:

**Algorithm 1:** Training the improved Sliced Wasserstein Generator

**Given** : Generator parameters $\theta_g$, Discriminator parameters $\theta_d, \omega_d$, sample size $n$, learning rate $\alpha$

**while** $\theta_g$ *not converged* **do**
  **for** $i \leftarrow 0$ **to** $k$ **do**
    Sample data $\{\mathcal{D}^i\}_{i=1}^n \sim \mathbb{P}_d$, generated samples $\{\mathcal{F}^i_{\theta_g}\}_{i=1}^n \sim \mathbb{P}_g$;
    compute **surogate loss** $s(\omega^T h_{\mathcal{D}}, \omega^T h_{\mathcal{F}(\theta_g)})$
      return $L \leftarrow s(\omega^T h_{\mathcal{D}}), \omega^T h_{\mathcal{F}(\theta_g)})$;
    $(\hat{\omega}, \hat{\theta}_d) \leftarrow (\hat{\omega}, \hat{\theta}_d) - \alpha \nabla_{\omega, \theta_d} L$;
  **end**
  compute **max-sliced Wasserstein Distance** *max-*$\tilde{W}_2(\hat{\omega}^T h_{\mathcal{D}}, \hat{\omega}^T h_{\mathcal{F}(\theta_g)})$
    Sample data $\{\mathcal{D}^i\}_{i=1}^n \sim \mathbb{P}_d$, generated samples $\{\mathcal{F}^i_{\theta_g}\}_{i=1}^n \sim \mathbb{P}_g$;
    sort $\hat{\omega}^T h_{\mathcal{D}}$ and $\hat{\omega}^T h_{\mathcal{F}(\theta_g)}$ to obtain permutations $\pi_{\mathcal{D}}, \pi_{\mathcal{F}}$;
    return $L = \sum_i \|\hat{\omega}^T h_{\mathcal{D}_{\pi_{\mathcal{D}}(i)}} - \hat{\omega}^T h_{\mathcal{F}_{\pi_{\mathcal{F}}(i)}(\theta_g)}\|_2^2$;
  $\theta_g \leftarrow \theta_g - \alpha \nabla_{\theta_g} L$;
**end**

**Definition 2** *Let $\Omega$ be the set of all directions on the unit sphere. Then, the max-sliced Wasserstein-2 distance between distributions $\mu$ and $\nu$ is defined as:*

$$\textit{max-}\tilde{W}_2(\mu, \nu) = \left[\max_{\omega \in \Omega} W_2^2(\mu^\omega, \nu^\omega)\right]^{\frac{1}{2}}. \qquad (13)$$

As illustrated in the following claim, it can be shown easily that max-$\tilde{W}_2(\cdot, \cdot)$ is a valid distance.

**Claim 2** *The max-sliced Wasserstein-2 distance defined in Eq. (13) is a well defined distance between distributions.*

**Proof.** See supplementary material. ■

We can also show that the max-sliced Wasserstein distance has polynomial sample complexity:

**Claim 3** *Consider the family of Gaussian distributions*

$$\mathcal{P} = \{\mathcal{N}(a, I) \mid a \in \mathbb{R}^d\}.$$

*The max-sliced Wasserstein-2 (max-$\tilde{W}_2$) distance is $\mathcal{P}$-generalizable.*

**Proof.** See the supplementary material. ■

Since it is a valid metric, we can directly use the max-sliced Wasserstein distance for learning distributions.

By definition, the max-sliced Wasserstein distance overcomes the limitation discussed in Sec. 3.2. However, we note that the use of a max-estimator is necessary, which is harder than estimation of a conventional random variable. In the following section, we discuss how the max-sliced Wasserstein distance can be estimated and used in a GAN-like setting.

### 3.4. max-sliced GAN

In this section, we discuss our approach that uses the max-sliced Wasserstein distance to train a GAN. We also discuss how we approximate the max-sliced Wasserstein distance in practice. Since we use max-$\tilde{W}_2$, we are able to achieve significant savings in terms of the number of projection directions needed as compared to [8].

Intuitively, we want to project data into a space where real samples can easily be differentiated from artificially generated points. To this end, we work with an adversarially learnt feature space, *i.e.*, we use the penultimate layer of a discriminator network. In this feature space, we minimize the max-sliced Wasserstein distance max-$\tilde{W}_2$. As will be discussed later in this section, finding the actual max is hard and therefore we resort to approximating it.

Let $\mathbb{P}_d$ again denote the data distribution and let $\mathbb{P}_g$ refer to the induced generator distribution. Further, let the discriminator be represented as $\omega_d^T h(.)$, where $\omega$ denotes the weights of a fully connected layer and $h$ represents the feature space we are interested in. Further, let $h_{\mathcal{D}}$ and $h_{\mathcal{F}}$ represent the two empirical distributions in this feature space. Then, we would like to solve

$$\text{max-}\tilde{W}_2(h_{\mathcal{D}}, h_{\mathcal{F}}) = \max_{\omega \in \Omega} W_2(h^\omega_{\mathcal{D}}, h^\omega_{\mathcal{F}}), \qquad (14)$$

where $\Omega$ is the set of all normalized directions. There is no easy way in general to solve

$$\omega^* = \operatorname*{argmax}_{\omega \in \Omega} W_2(h^\omega_{\mathcal{D}}, h^\omega_{\mathcal{F}}), \qquad (15)$$

even if the parameters $\theta_d$ of the feature transform $h$ are fixed. This is because computation of the Wasserstein distance $W_2(h^\omega_{\mathcal{D}}, h^\omega_{\mathcal{F}})$ in the 1-dimensional case requires sorting, *i.e.*, solving of a minimization problem. Hence the program given in Eq. (15) is a saddlepoint objective, for which both maximization and minimization can be solved exactly when assuming the parameters of the other program to be fixed.

| | en-es | es-en | en-fr | fr-en | en-de | de-en | en-ru | ru-en | en-zh | zh-en |
|---|---|---|---|---|---|---|---|---|---|---|
| [6] - NN | 79.1 | 78.1 | 78.1 | 78.2 | 71.3 | 69.6 | 37.3 | 54.3 | 30.9 | 21.9 |
| [6] - CSLS | 81.7 | 83.3 | 82.3 | 82.1 | 74.0 | 72.2 | 44.0 | 59.1 | 32.5 | 31.4 |
| Max-sliced WGAN - NN | 79.6 | 79.1 | 78.2 | 78.5 | 71.9 | 69.6 | 38.4 | 58.7 | 34.9 | 25.1 |
| Max-sliced WGAN - CSLS | **82.0** | **84.1** | **82.5** | **82.3** | **74.8** | **73.1** | **44.6** | **61.7** | **35.3** | **31.9** |

Table 1: Unsupervised word translation. We show the retrieval precision P@1 on 5 pairs of languages on MUSE bilingual dictionaries [6]: English ('en'), French ('fr'), German ('de'), Russian ('ru') and Chinese ('zh').

If we want to jointly find the parameters $\theta_d$ of the feature transform $h$ and the projection direction $\omega$, *i.e.*, if we want to solve

$$\omega^*, \theta_d^* = \operatorname*{argmax}_{\omega\in\Omega,\theta_d} W_2(h_{\mathcal{D}}^{\omega}, h_{\mathcal{F}}^{\omega}), \tag{16}$$

using gradient descent based methods, we also need to pay attention to bounded-ness of the objective. Using regularization often proves tricky and may require separate tuning for each use case.

To circumvent those difficulties when jointly searching for $\omega^*$ and $\theta_d^*$, we use a surrogate function $s$ and write the objective for the discriminator as follows:

$$\hat{\omega}, \hat{\theta}_d = \operatorname*{argmax}_{\omega\in\Omega,\theta_d} s(\omega^T h_{\mathcal{D}}, \omega^T h_{\mathcal{F}}). \tag{17}$$

Intuitively, and in spirit similar to max-$\tilde{W}_2$, we want the surrogate function $s$ to transform the data via $h$ into a space where $h_{\mathcal{D}}$ and $h_{\mathcal{F}}$ are easy to differentiate. Moreover, we want $\omega$ to be the direction which best separates the transformed real and generated data. A variety of surrogate functions such as the log-loss as specified in Eq. (8), the hinge-loss, or a moment separator with

$$s(\omega^T h_{\mathcal{D}}, \omega^T h_{\mathcal{F}}) = \sum_{x\in\mathcal{D}} \omega^T h_x - \sum_{\hat{x}\in\mathcal{F}} \omega^T h_{\hat{x}} \tag{18}$$

come to mind immediately.

For instance, in case of a log-loss, $\omega^T h$ learns to classify real and fake samples, essentially performing linear logistic regression using $\omega$ on a learned feature representation $h$. If trained to optimality, the two distributions are well separated in the discriminator's feature space $h$. An example is given in Fig. 2. The discriminator takes two distributions, shown in Fig. 2a and is trained to classify them. In doing so the discriminator transforms them to the feature space shown in Fig. 2b. In this simple example, we can plot the Wasserstein distance along the different projection directions. This is visualized in Fig. 2c. The discriminator's final layer can be considered as a projection direction. This direction is very close to the maximizer of the projected Wasserstein distance in the feature space.

Additionally, in this case, $\omega^*$ can be approximated with $\hat{\omega}$ – because the discriminator, trained for classification, essentially separates the distributions along $\hat{\omega}$. If we compute the Wasserstein-2 distance for projections onto different angles (as in Fig. 2c), we see that the maximum distance is achieved close to the projection direction from the discriminator, *i.e.*, $\hat{\omega}$. We next assess: 'how close?'

While log-loss and all other functions seem intuitive, we provide for the special case of the moment separator given in Eq. (18) and an identity transform $h$ the maximal suboptimality in terms of the max-sliced Wasserstein distance:

**Claim 4** *For the surrogate function $s$ given in Eq. (18), $h$ the identity, and $\hat{\omega}$ computed as specified in Eq. (17), we obtain*

$$\alpha(\mathcal{D},\mathcal{F}) \le W_2^2(\mathcal{D}^{\hat{\omega}}, \mathcal{F}^{\hat{\omega}}) \le V^* = \textit{max-}\tilde{W}_2(\mathcal{D},\mathcal{F})^2,$$

*for a lower bound $\alpha(\mathcal{D},\mathcal{F}) = \|m\|_2^2$, where $m = \sum_i \mathcal{D}_i - \sum_i \mathcal{F}_i$ is the difference of dataset means.*

**Proof.** See the supplementary material. ■

To summarize, training the discriminator for classification provides a rich feature space which can be utilized for faster training. We note that the discriminator might be trained to obtain such features in a more explicit manner, but we leave this to future research.

### 3.5. max-sliced GAN Algorithm

We summarize the resulting training process in Alg. 1. It proceeds as follows: In every iteration, we draw a set of samples $\mathcal{D}$ and $\mathcal{F}$ from the true and fake distributions. We optimize the parameters $\theta_d$ and $\omega$ of the feature transform $h$ for $k$ iterations ($k$ is a hyper-parameter) to maximize a surrogate loss function $s(\omega^T h_{\mathcal{D}}, \omega^T h_{\mathcal{F}})$. Then we compute the Wasserstein-2 distance between the output distributions of the discriminator, *i.e.*, $W_2(\hat{\omega}^T h_{\mathcal{D}}, \hat{\omega}^T h_{\mathcal{F}})$. The generator is trained to minimize this distance. In our experiments, we choose $h$ to be the binary classification loss.

## 4. Experiments

In this section, we present results to demonstrate the effectiveness of the max-sliced Wasserstein distance and the computational benefits it offers over the sliced Wasserstein distance. We show quantitative results on unpaired word translation [6], and qualitative and quantitative results on image generation tasks using the CelebA-HQ [16] and the LSUN Bedrooms [36] datasets.

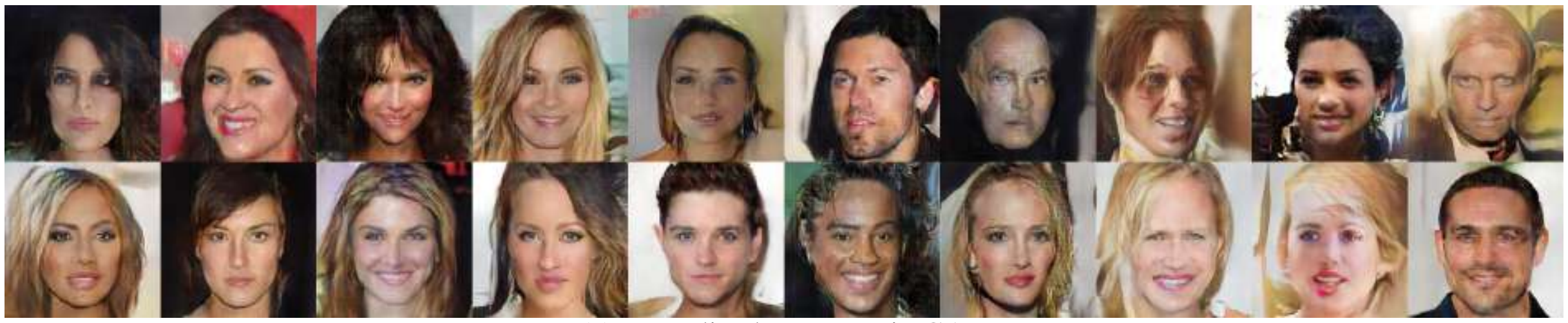

(a) Max-sliced Wasserstein GAN

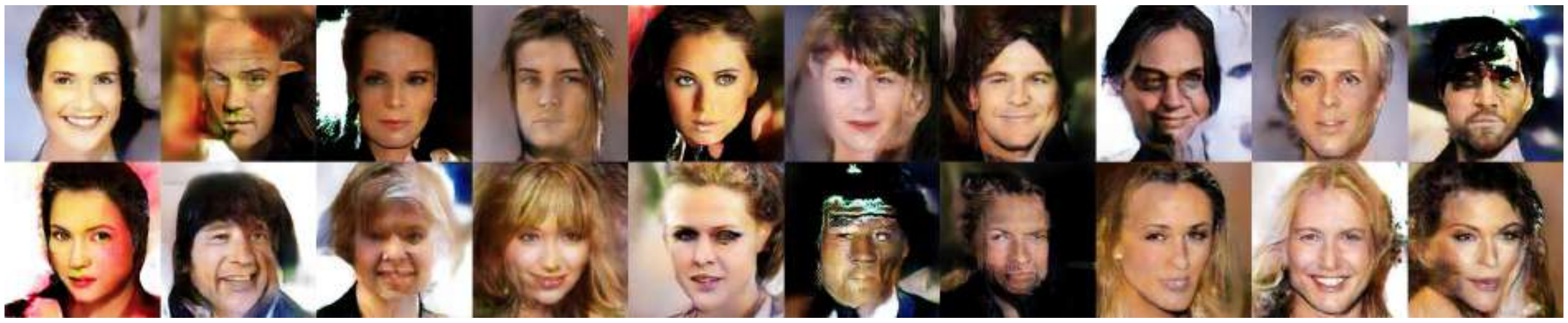

(b) Sliced Wasserstein GAN with 100 projections

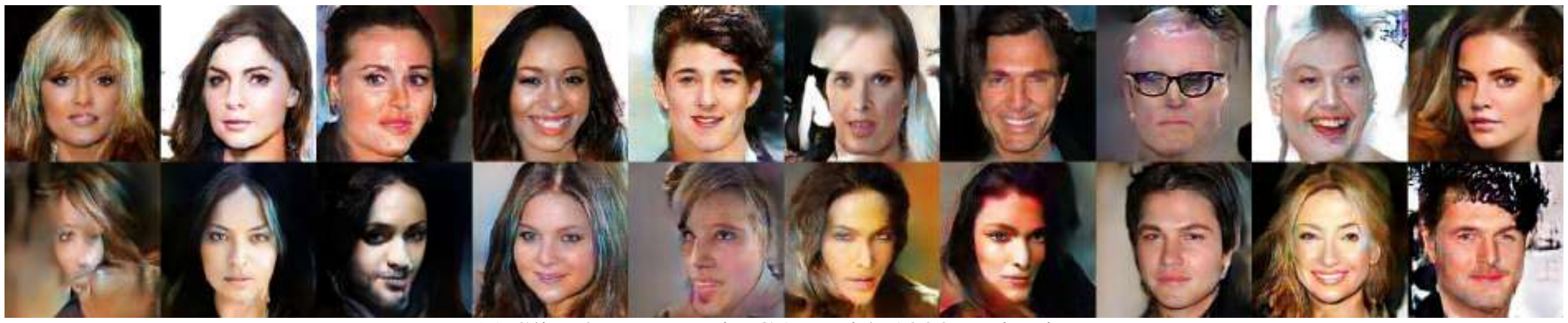

(c) Sliced Wasserstein GAN with 1000 projections

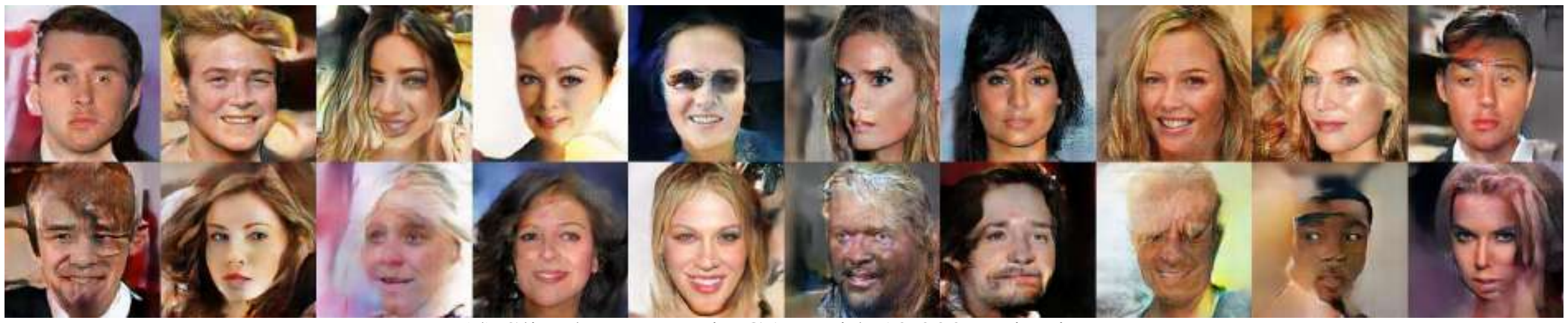

(d) Sliced Wasserstein GAN with 10,000 projections

Figure 3: Generated samples ($256 \times 256$) from CelebA-HQ.

### 4.1. Word Translation without Parallel Data

We evaluate the effectiveness of the max-sliced GAN on unsupervised word translation tasks, *i.e.*, without paired/parallel data [6]. This allows us to quantitatively compare different methods.

The setting of this experiment is as follows. We are given embeddings of words from two languages, say $X, Y \in \mathbb{R}^d$. We want to learn an orthogonal transformation $W^*$ that maps the source embeddings $X$ to $Y$, *i.e.*:

$$W^* = \underset{W \in \mathbb{R}^{d \times d}, \text{orthogonal}}{\operatorname{argmin}} ||WX - Y||_F. \tag{19}$$

The current state-of-the-art [6] employs a GAN-like [11] adversary to learn the transformation. Therefore, the transformation is learned by minimizing the Jenson-Shannon divergence between $WX$ and $Y$. We instead minimize the max-sliced Wasserstein distance to learn $W$.

We follow the training method and evaluation in [6] and report the word translation precision by computing the retrieval precision@k for $k = 1$ on the MUSE bilingual dictionaries [6]. During testing, 1,500 queries are tested and 200k words of the target language are taken into account. We compare our method with [6] and present results for 5 pairs of languages in Tab. 1. In Tab. 1 'NN' represents use of nearest neighbors to build the dictionary after training the transformation $W$, and 'CSLS' stands for use of cross-domain similarity local scaling [6]. Our method with CSLS outperforms the baseline in all tested language pairs. This demonstrates the competitiveness of our method with current established GAN frameworks.

### 4.2. Image Generation

In this section, we present results on the task of image generation. Using the max-sliced Wasserstein distance, we train a GAN on the CelebA [16] and LSUN Bedrooms [36] datasets for images of resolution 256x256. We compare with the sliced Wasserstein GAN [8].

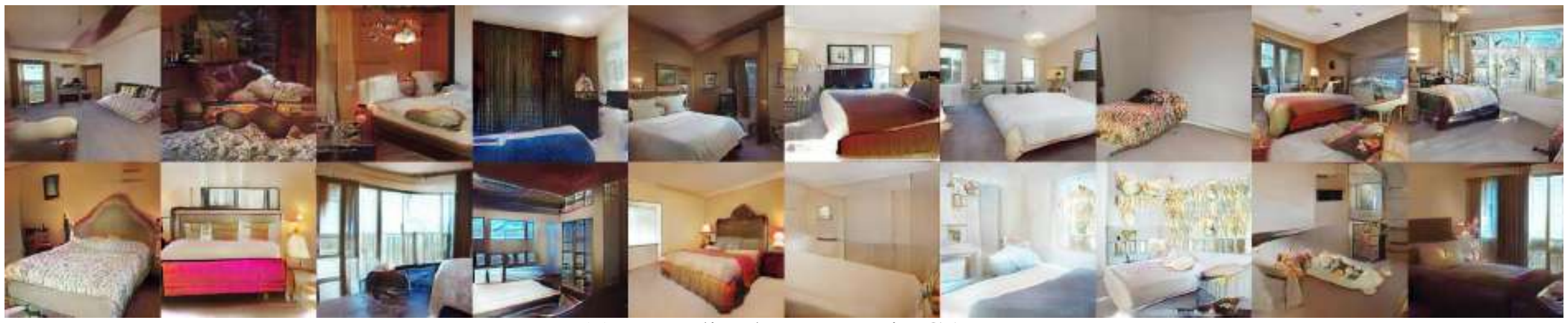
(a) Max-sliced Wasserstein GAN

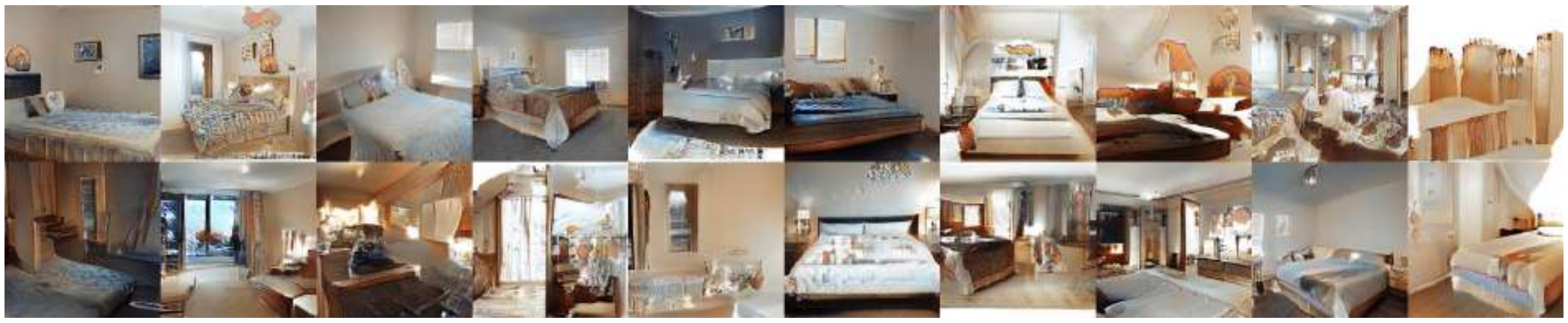
(b) Sliced Wasserstein GAN with 100 projections

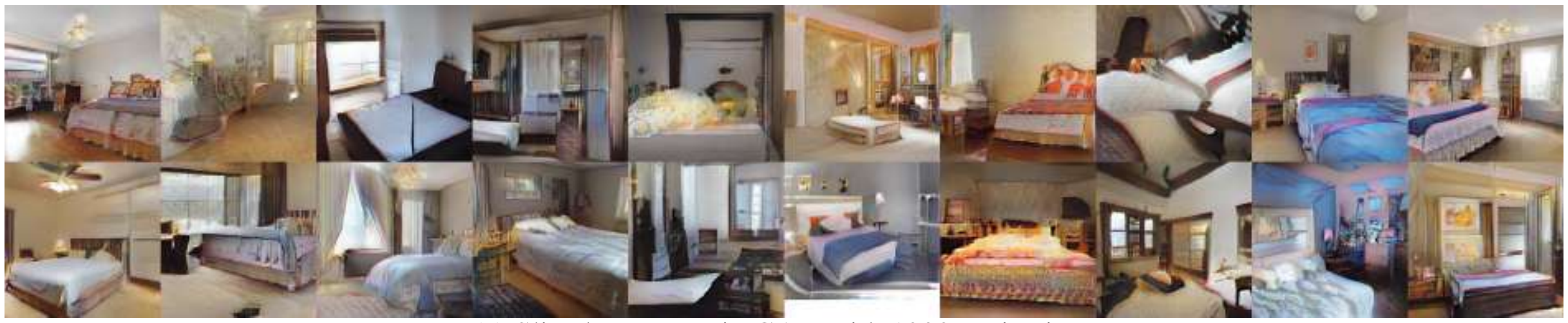
(c) Sliced Wasserstein GAN with 1000 projections

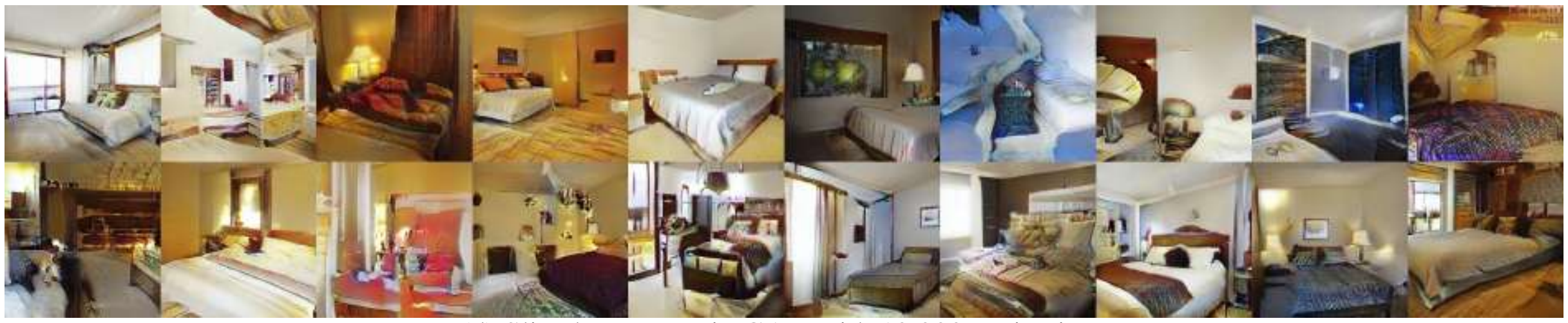
(d) Sliced Wasserstein GAN with 10,000 projections

Figure 4: Generated samples ($256 \times 256$) from LSUN Bedrooms.

Samples generated by each trained model are presented in Fig. 3 and Fig. 4. The results of the max-sliced Wasserstein GAN are shown Fig. 3a and Fig. 4a. We train the sliced Wasserstein GAN with 100, 1000, and 10000 random projections. Results of each of these are respectively shown in Fig. 3b, Fig. 3c, and Fig. 3d for CelebA-HQ, and in Fig. 4b, Fig. 4c, and Fig. 4d for LSUN. The max-sliced Wasserstein GAN using just one projection direction is able to produce results which are either comparable or better than the sliced Wasserstein GAN even when using 10000 projections. This significantly reduces the computational complexity and also the memory footprint of the model.

We used a simple extension of the popular DCGAN architecture for the generator and discriminator. Two extra strided (transpose) convolutional layers are added to the generator and the discriminator to scale to 256x256. We do not use any special normalization/ initialization to train the models. Specific details are given in the supplementary.

## 5. Conclusion

In this paper, we analyzed the Wasserstein and sliced Wasserstein distance and developed a simple yet effective training strategy for generative adversarial nets based on the max-sliced Wasserstein distance. We showed that this distance enjoys a better sample complexity than the Wasserstein distance, and a better projection complexity than the sliced Wasserstein distance. We developed a method to approximate it using a surrogate loss, and also analyzed the approximation error for one such surrogate. Empirically, we showed that the discussed approach is able to learn high dimensional distributions. The method requires orders of magnitude fewer projection directions than the sliced Wasserstein GAN even though both work in a similar distance space.

**Acknowledgments:** This work is supported in part by NSF under Grant No. 1718221, Samsung, and 3M. We thank NVIDIA for providing GPUs used for this work.

## References

[1] M. Arjovsky and L. Bottou. Towards principled methods for training generative adversarial networks. In *ICLR*, 2017. 2

[2] M. Arjovsky, S. Chintala, and L. Bottou. Wasserstein gan. In *ICML*, 2017. 1, 2

[3] S. Arora, R. Ge, Y. Liang, T. Ma, and Y. Zhang. Generalization and equilibrium in generative adversarial nets (gans). In *ICML*, 2017. 1

[4] D. Berthelot, T. Schumm, and L. Metz. Began: Boundary equilibrium generative adversarial networks. *arXiv preprint arXiv:1703.10717*, 2017. 1

[5] N. Bonneel, J. Rabin, G. Peyré, and H. Pfister. Sliced and radon wasserstein barycenters of measures. *Journal of Mathematical Imaging and Vision*, 2015. 2

[6] A. Conneau, G. Lample, M. Ranzato, L. Denoyer, and H. Jegou. Word translation without parallel data. In *ICLR*, 2018. 2, 6, 7

[7] R. W. A. Cully, H. J. Chang, and Y. Demiris. Magan: Margin adaptation for generative adversarial networks. *arXiv preprint arXiv:1704.03817*, 2017. 1

[8] I. Deshpande, Z. Zhang, and A. Schwing. Generative modeling using the sliced wasserstein distance. In *CVPR*, 2018. 1, 2, 5, 7

[9] J. Donahue, P. Krähenbühl, and T. Darrell. Adversarial feature learning. In *ICLR*, 2017. 1

[10] S. Feizi, C. Suh, F. Xia, and D. Tse. Understanding gans: the lqg setting. *arXiv preprint arXiv:1710.10793*, 2017. 1

[11] I. Goodfellow, J. Pouget-Abadie, M. Mirza, B. Xu, D. Warde-Farley, S. Ozair, A. Courville, and Y. Bengio. Generative adversarial nets. In *NIPS*, 2014. 1, 7

[12] I. Gulrajani, F. Ahmed, M. Arjovsky, V. Dumoulin, and A. Courville. Improved training of wasserstein gans. In *NIPS*, 2017. 1, 2

[13] M. Heusel, H. Ramsauer, T. Unterthiner, B. Nessler, and S. Hochreiter. Gans trained by a two time-scale update rule converge to a local nash equilibrium. In *NIPS*, 2017. 1

[14] X. Huang, M.-Y. Liu, S. Belongie, and J. Kautz. Multimodal Unsupervised Image-to-Image Translation. In *Proc. ECCV*, 2018. 1

[15] P. Isola, J.-Y. Zhu, T. Zhou, and A. A. Efros. Image-to-image translation with conditional adversarial networks. In *CVPR*, 2017. 1

[16] T. Karras, T. Aila, S. Laine, and J. Lehtinen. Progressive growing of gans for improved quality, stability, and variation. In *ICLR*, 2017. 1, 2, 6, 7

[17] D. P. Kingma and M. Welling. Auto-encoding variational bayes. *arXiv preprint arXiv:1312.6114*, 2013. 1

[18] S. Kolouri, C. E. Martin, and G. K. Rohde. Sliced-wasserstein autoencoder: An embarrassingly simple generative model. *arXiv preprint arXiv:1804.01947*, 2018. 2

[19] S. Kolouri, S. R. Park, and G. K. Rohde. The radon cumulative distribution transform and its application to image classification. *IEEE transactions on image processing*, 2016. 2

[20] S. Kolouri, G. K. Rohde, and H. Hoffman. Sliced wasserstein distance for learning gaussian mixture models. In *CVPR*, 2018. 1, 2

[21] H. Y. Lee, H. Y. Tseng, J. B. Huang, M. K. Singh, and M. H. Yang. Diverse image-to-image translation via disentangled representation. In *Proc. ECCV*, 2018. 1

[22] C.-L. Li, W.-C. Chang, Y. Cheng, Y. Yang, and B. Póczos. Mmd gan: Towards deeper understanding of moment matching network. In *NIPS*, 2017. 1

[23] Z. Lin, A. Khetan, G. Fanti, and S. Oh. Pacgan: The power of two samples in generative adversarial networks. In *NIPS*, 2018. 1

[24] M.-Y. Liu, T. Breuel, and J. Kautz. Unsupervised Image-to-Image Translation Networks. In *Proc. NIPS*, 2017. 1

[25] L. Mescheder, A. Geiger, and S. Nowozin. Which training methods for gans do actually converge? In *ICML*, 2018. 1

[26] Y. Mroueh and T. Sercu. Fisher gan. In *NIPS*, 2017. 1

[27] Y. Mroueh, T. Sercu, and V. Goel. Mcgan: Mean and covariance feature matching gan. *arXiv preprint arXiv:1702.08398*, 2017. 1

[28] A. Radford, L. Metz, and S. Chintala. Unsupervised representation learning with deep convolutional generative adversarial networks. In *ICLR*, 2016. 1

[29] A. Royer, K. Bousmalis, S. Gouws, F. Bertsch, I. Moressi, F. Cole, and K. Murphy. Xgan: Unsupervised image-to-image translation for many-to-many mappings. In *arXiv:1711.05139*, 2017. 1

[30] T. Salimans, H. Zhang, A. Radford, and D. Metaxas. Improving gans using optimal transport. In *ICLR*, 2018. 1

[31] A. Shrivastava, T. Pfister, O. Tuzel, J. Susskind, W. Wang, and R. Webb. Learning from simulated and unsupervised images through adversarial training. In *CVPR*, 2017. 1

[32] C. Villani. *Optimal transport: old and new*. Springer Science & Business Media, 2008. 2

[33] J. Weed and F. Bach. Sharp asymptotic and finite-sample rates of convergence of empirical measures in wasserstein distance. *arXiv preprint arXiv:1707.00087*, 2017. 2

[34] J. Wu, Z. Huang, W. Li, J. Thoma, and L. Van Gool. Sliced wasserstein generative models. *arXiv preprint arXiv:1706.02631*, 2017. 1, 2

[35] Z. Yi, H. Zhang, P. Tan, and M. Gong. Dualgan: Unsupervised dual learning for image-to-image translation. In *Proc. ICCV*, 2017. 1

[36] F. Yu, A. Seff, Y. Zhang, S. Song, T. Funkhouser, and J. Xiao. Lsun: Construction of a large-scale image dataset using deep learning with humans in the loop. *arXiv preprint arXiv:1506.03365*, 2015. 2, 6, 7

[37] J.-Y. Zhu, T. Park, P. Isola, and A. A. Efros. Unpaired image-to-image translation using cycle-consistent adversarial networks. In *ICCV*, 2017. 1

[38] J. Y. Zhu, R. Zhang, D. Pathak, T. Darrell, A. A. Efros, O. Wang, and E. Shechtman. Toward multimodal image-to-image translation. In *Proc. NIPS*, 2017. 1

# Supplementary Material: Max-Sliced Wasserstein Distance and its use for GANs

Ishan Deshpande, Yuan-Ting Hu, Ruoyu Sun, Ayis Pyrros[†], Nasir Siddiqui[†],
Sanmi Koyejo, Zhizhen Zhao, David Forsyth, Alexander Schwing
University of Illinois at Urbana-Champaign [†]Dupage Medical Group
ishan.sd@gmail.com, {ythu2, ruoyus}@illinois.edu, ayis@ayis.org, nsiddiqui@gmail.com,
{sanmi, zhizhenz, daf, aschwing}@illinois.edu

## 1. Experiments with Images at Higher Resolutions

In this section we repeat the experiments presented in the paper at an increased resolution of 512x512. We first compare the max-sliced Wasserstein GAN to the sliced Wasserstein GAN (with 100, 1000 and 10,000 projections) after 50,000 iterations of training in Fig. 1. This reiterates and validates the claims of the paper that the max-sliced Wasserstein GAN provides faster convergence than the sliced Wasserstein GAN with much fewer projections, and this is noticeable especially during the *early stages* of training. The difference is more pronounced as the distribution dimensions increase (for instance, the benefit is more obvious when training on images at a resolution of 512x512 compare to 256x256).

We then present random samples from the max-sliced Wasserstein GAN in Fig. 2. Generated images are shown when the input noise is sampled from the original sampling distribution $\mathcal{N}(0, I_{128})$, as well as from a scaled version, *i.e.*, $0.05 \times \mathcal{N}(0, I_{128})$. The model is trained end-to-end in a single unified process, *i.e.*, no progressive growing, stacking, or any other tricks and is a simple architecture (as described in Sec. 4).

## 2. Proof of Claim 1 and Claim 3 in the paper

In this part of the supplementary material, we prove Claim 1 and Claim 3 of the paper. The claims state that the max-sliced Wasserstein distance and the sliced Wasserstein distance (Gaussian version) are $\mathcal{P}$-generalizable for a class of Gaussian distributions defined as

$$\mathcal{P} = \{\mathcal{N}(a, I_d) \mid a \in \mathbb{R}^d\},$$

while the Wasserstein distance is not. We restate the main result as below.

**Claim 1** *(combination of Claim 1 and Claim 3 in the paper) We say the distance* $dist(\cdot,\cdot)$ *is* $\mathcal{P}$*-generalizable if it satisfies the following: for any two random distributions* $\mu$, $\nu$ *from the family* $\mathcal{P}$ *and their empirical versions* $\hat{\mu}$, $\hat{\nu}$ *each with* $n = poly(d, 1/\epsilon)$ *samples (here* $poly(d, 1/\epsilon)$ *means a certain polynomial of* $d, 1/\epsilon$*), with high probability* [1] *we have*

$$Pr(|dist(\mu, \nu) - dist(\hat{\mu}, \hat{\nu})| \leq \epsilon). \quad (1)$$

*Consider the family of distributions* $\mathcal{P} = \{\mathcal{N}(a, I_d) \mid a \in \mathbb{R}^d\}$*. The max-sliced Wasserstein-2 distance and the sliced Wasserstein distance (Gaussian version) are* $\mathcal{P}$*-generalizable, while the Wasserstein distance is not.*

**Proof**: Without loss of generality, we assume $\nu \sim \mathcal{N}(0, I_d)$, $\mu \sim \mathcal{N}(\beta e_1, I_d)$ and $\beta \geq 0$. This is because the Gaussian distribution in the family $\mathcal{P}$ is isotropic. Hence we can always rotate the two distributions such that the mean of the two new distributions differ only in the first coordinate. Suppose $\hat{\mu}$ consists of vectors $x^1, \ldots, x^n \in \mathbb{R}^d$, and $\hat{\nu}$ consists of vectors $y^1, \ldots, y^n \in \mathbb{R}^d$.

In the following three sections, we analyze the Wasserstein distance, the max-sliced Wasserstein distance and the sliced Wasserstein distance separately.

[1]The probability is taken with respect to the choice of the samples. The exact probability will be specified in the results.

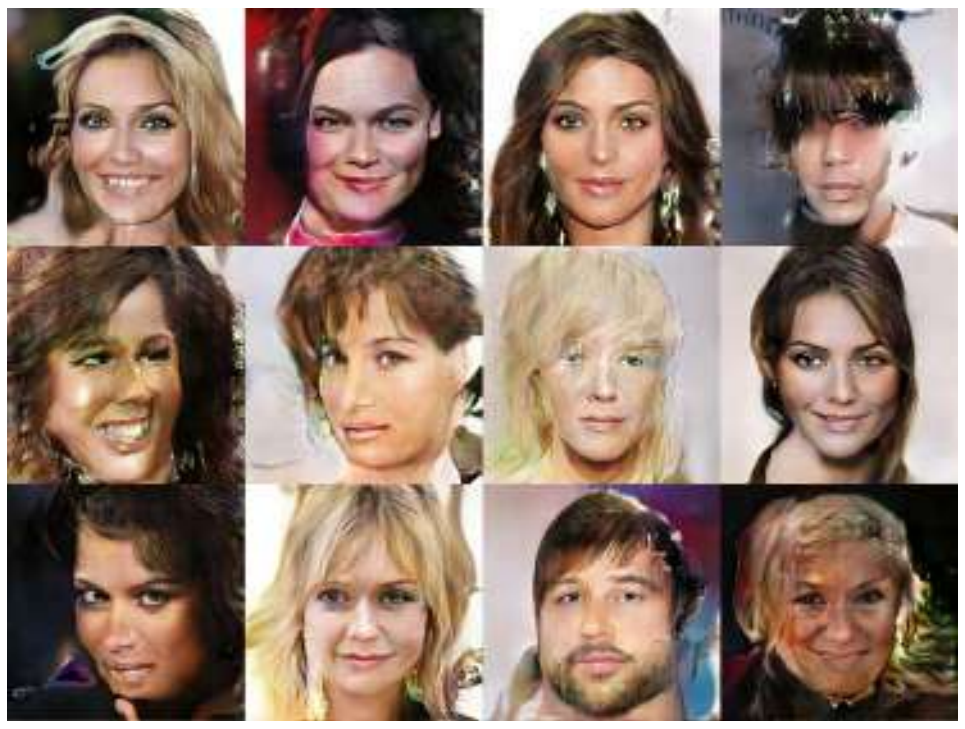

(a) Max-sliced Wasserstein GAN

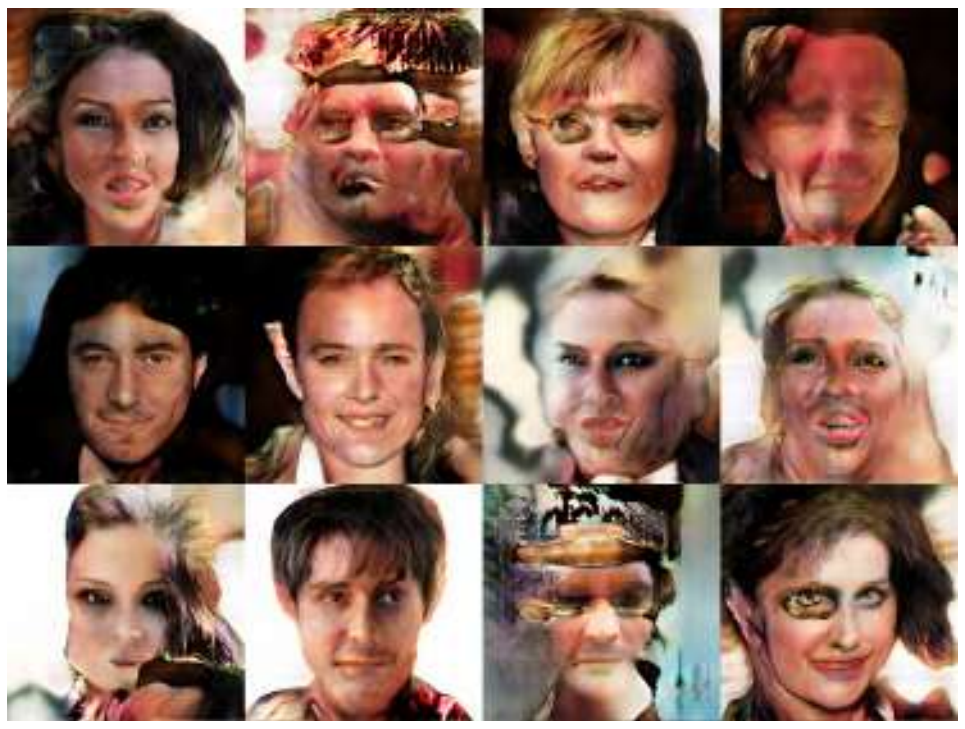

(b) Sliced Wasserstein GAN with 100 projections

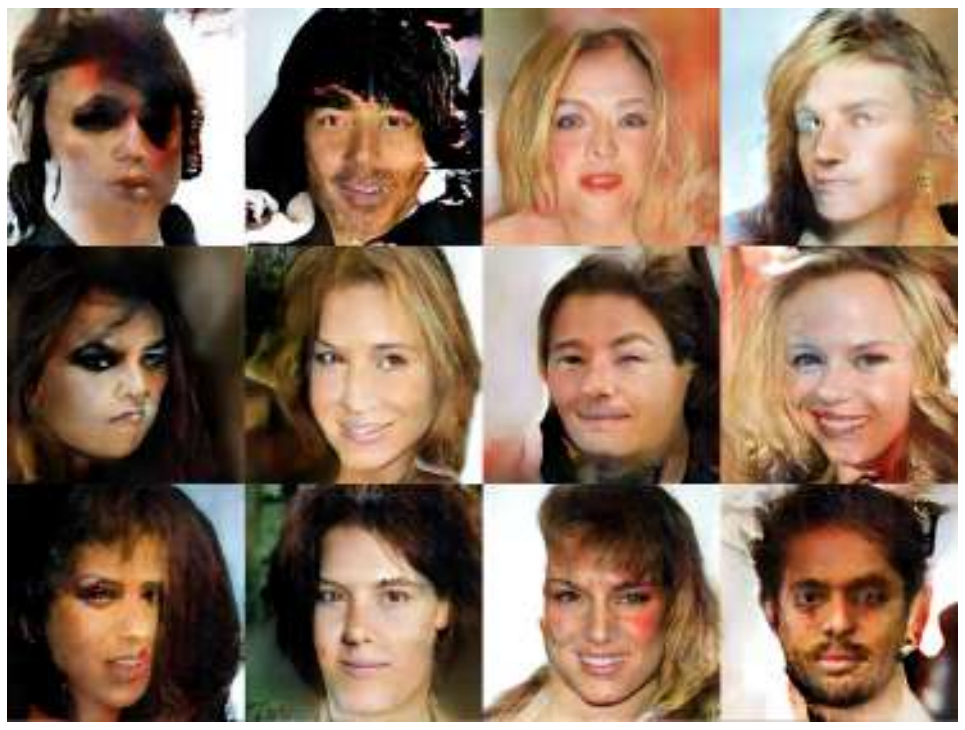

(c) Sliced Wasserstein GAN with 1000 projections

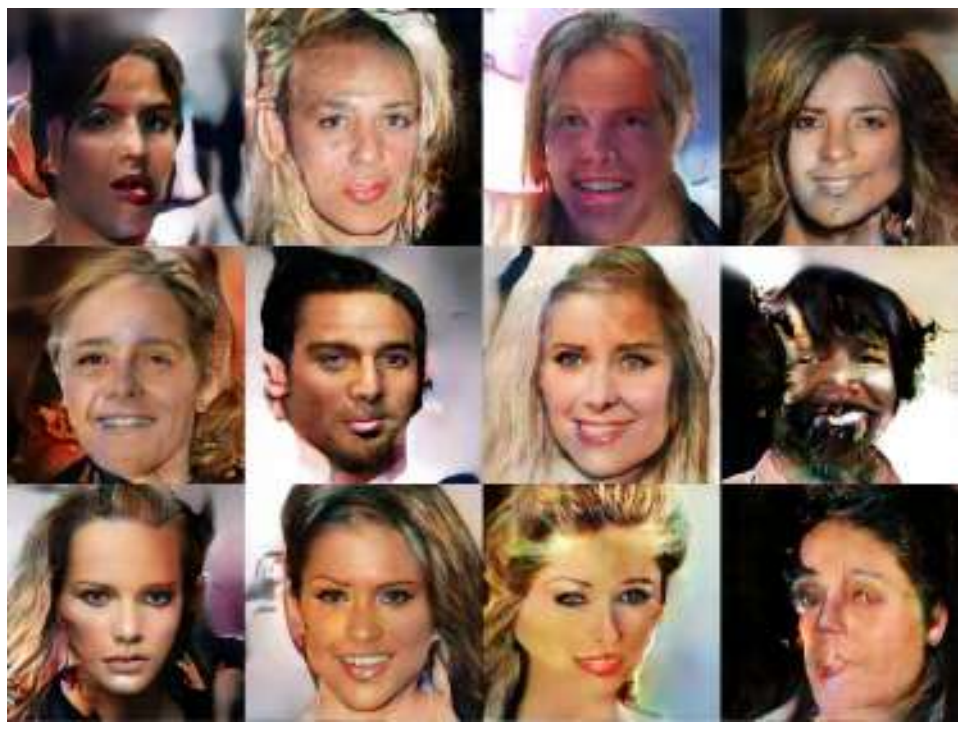

(d) Sliced Wasserstein GAN with 10,000 projections

Figure 1: Random samples (512x512) after 50,000 iterations of training on CelebA-HQ.

## 2.1. Wasserstein distance is not $\mathcal{P}$-generalizable

We show that with a polynomial number of samples, the empirical Wasserstein distance is not a good approximation of the true Wasserstein distance. We will only consider the special case $\beta = 0$. In this case, the two distributions $\mu, \nu$ are both $\mathcal{N}(0, I)$, thus the population Wasserstein distance is 0.

For any given $i, j$, $\|x^i - y^j\|^2$ is the squared sum of $d$ independent Gaussian variables, each with variance 2. Hence, $\|x^i - y^j\|^2$ follows a Chi-square distribution with variance $2d$. From standard tail bounds (e.g. [4, Lemma 1]), we have that for any $t > 0$,

$$Pr(\|x^i - y^j\|^2/2 - d \leq -2\sqrt{d}t) \leq \exp(-t^2).$$

Let $t = \sqrt{d}/4$, then the above relation implies that with probability at least $1 - \exp(-d/16)$, we have $\|x^i - y^j\|^2 \geq d$. Using the union bound, with probability at least $1 - n^2 \exp(d/16)$, the $n^2$ distances $\|x^i - y^j\|^2 \; \forall i, j \in \{1, \dots, n\}$ are larger than $d$, then the empirical Wasserstein distance $W_2(\mu, \nu) > \sqrt{d}$. When $n = \text{poly}(d)$, the probability $1 - n^2 \exp(-d/16) = 1 - \text{poly}(d) \exp(-d/16)$ is larger than, say, $1 - \exp(d/32)$ for large enough $d$. Thus for large enough $d$, with high probability, the empirical Wasserstein distance $W_2(\hat{\mu}, \hat{\nu}) > \sqrt{d}$, which is much larger than the true distance 0. Thus the Wasserstein distance is not $\mathcal{P}$-generalizable. ■

**Remark**: In the case of exponentially many samples, *i.e.*, $n = e^{\Omega(d)}$, the probability $1 - n^2 \exp(-d/16)$ can be small, and the above argument does not hold.

## 2.2. Proof of max-sliced Wasserstein distance is $\mathcal{P}$-generalizable

We formally state the result that the max-sliced Wasserstein distance is $\mathcal{P}$-generalizable.

**Proposition 2.1** *The max-sliced Wasserstein distance is $\mathcal{P}$-generalizable in the following sense. Suppose $\mu$ and $\nu$ are two distributions of the family $\mathcal{P}$, and $\hat{\mu}, \hat{\nu}$ are two samples of $\mu, \nu$ each with size $n$. There exist some numerical constants $C, C'$,*

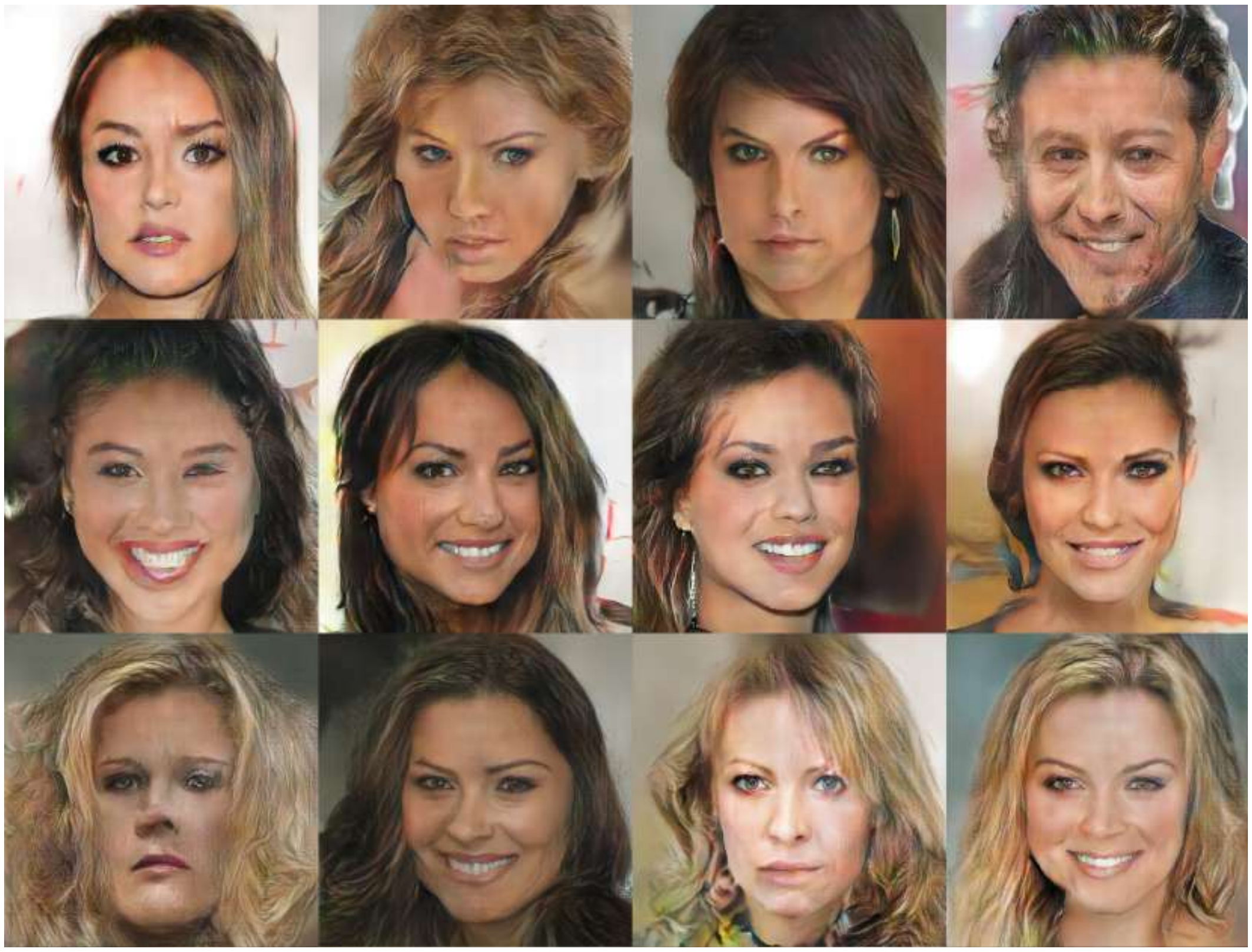

(a) Max-sliced Wasserstein GAN, noise from $\mathcal{N}(0, I)$

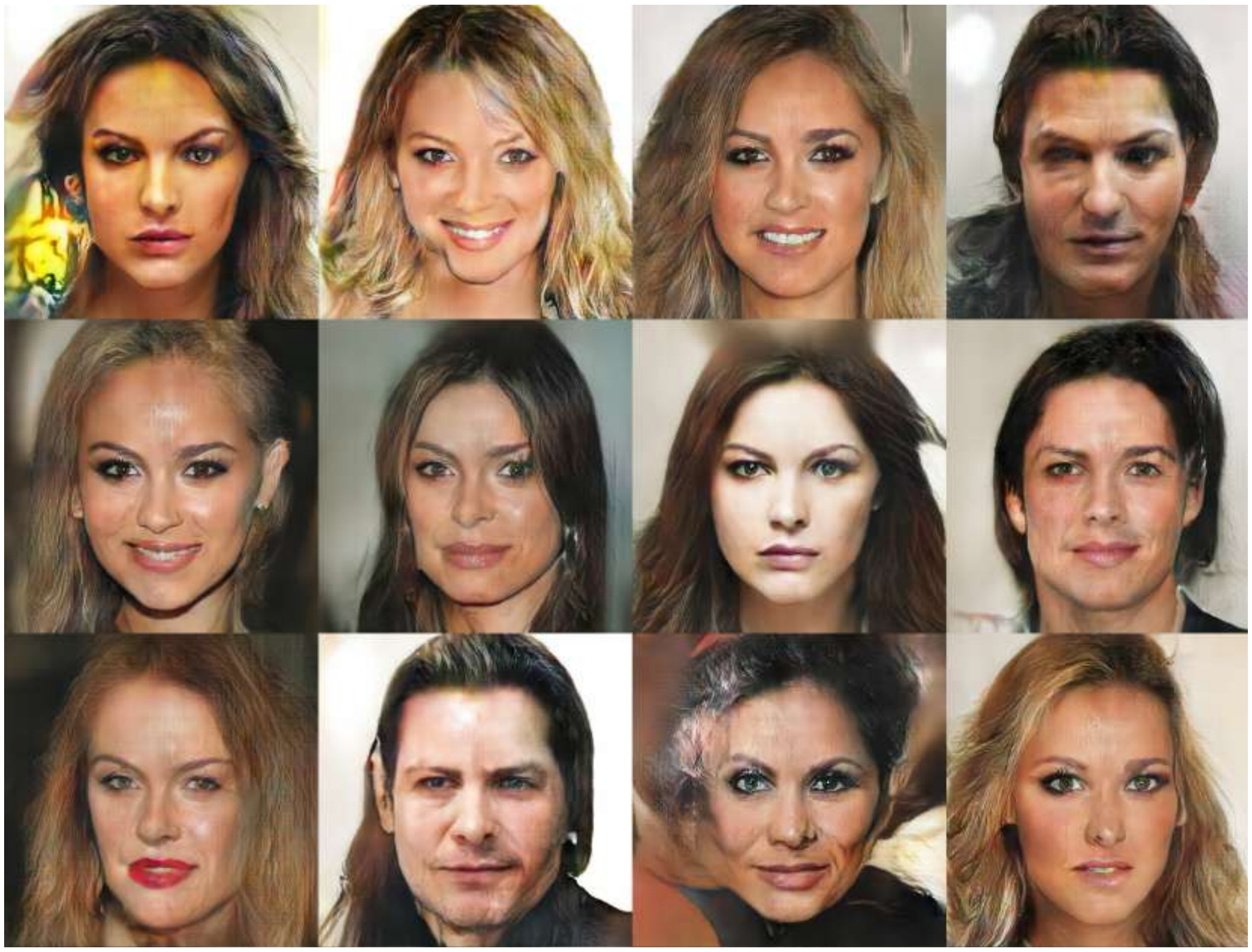

(b) Max-sliced Wasserstein GAN, noise from $0.05\mathcal{N}(0, I)$

Figure 2: Random samples (512x512) from the max-sliced Wasserstein GAN trained on CelebA-HQ.

*such that for any $m \geq 0$ and any $0 < \epsilon < 1/C'$, when*

$$n \geq C \frac{1}{\epsilon^2}(d \log \frac{1}{\epsilon} + m),$$

*with probability at least $1 - 5n^{-8} - \exp(-m)$, we have*

$$|\textit{max-}\tilde{W}_2(\mu, \nu) - \textit{max-}\tilde{W}_2(\hat{\mu}, \hat{\nu})| \leq \epsilon.$$

**Proof:** As mentioned earlier, without loss of generality, we assume $\nu \sim \mathcal{N}(0, I_d)$ and $\mu \sim \mathcal{N}(\beta e_1, I_d)$. For any unit vector $\omega \in \mathbb{R}^d$, we have $\omega^T \mu \sim \mathcal{N}(\beta\omega_1, 1)$ and $\omega^T \nu \sim \mathcal{N}(0, 1)$. The Wasserstein-2 distance between the two distributions is

$$W_2(\omega^T \mu, \omega^T \nu) = \|\beta\omega_1 - 0\| = \beta|\omega_1|.$$

Therefore

$$\max_{\|\omega\|=1} W_2(\omega^T \mu, \omega^T \nu) = \max_{\|\omega\|=1} \beta|\omega_1| = \beta,$$

and the equality is achieved when $\omega = e_1$. In other words, the optimal projection direction is the one that connects the center of the two Gaussian distributions. This also shows that the population max-sliced Wasserstein distance is $\beta$.

We then show that with a polynomial number of samples the max-sliced Wasserstein distance $\beta$ can be well approximated. For a given direction $\omega \in \mathbb{R}^d$, we project $x^i, y^j$ onto this direction to get the $2n$ 1-dimensional samples $\omega^T x^1, \dots, \omega^T x^n$, $\omega^T y^1, \dots, \omega^T y^n$. Define two vectors $\hat{x} = (\omega^T x^1, \dots, \omega^T x^n)$ and $\hat{y} = (\omega^T y^1, \dots, \omega^T y^n)$. Now the Wasserstein-2 distance along this direction is given by

$$W_2(\hat{\mu}^\omega, \hat{\nu}^\omega) = \sqrt{\frac{1}{n}\sum_i (\hat{x}_{[i]} - \hat{y}_{[i]})^2},$$

where $z_{[1]} \geq z_{[2]} \geq \cdots \geq z_{[n]}$ denotes the sorted elements of any vector $z$.

The problem of finding the maximal Wasserstein-2 distance is stated as follows:

$$\max_{\|\omega\|=1} \phi(\omega) \triangleq \frac{1}{n}\sum_i (\hat{x}_{[i]} - \hat{y}_{[i]})^2.$$

Denote

$$v^* = \max_{\|\omega\|=1} \phi(\omega).$$

Obviously, $v^* = \text{max-}\tilde{W}_2(\hat{\mu}, \hat{\nu})^2$, as defined in Eq. (13) of the main paper. It is not clear how to obtain an analytical expression of the optimal value. Nevertheless, for our purpose we only need to give an estimate of this objective value $v^*$. We will first present an estimate of $\phi(e_1)$, which will give a lower bound of $v^*$ (since $v^*$ is the optimal value). Then we prove that $v^*$ is upper bounded by the square of $\beta$ plus some small error. Together they imply that the empirical max-sliced Wasserstein distance $\sqrt{v^*}$ is close to the population max-sliced Wasserstein distance $\beta$.

**Lemma 2.1** *There exists a numerical constant $C_2$ such that with probability at least $1 - 2n^{-8}$, we have*

$$\beta^2 - \sqrt{\frac{\log n}{n}} 8\beta \leq \phi(e_1) \leq \beta^2 + C_2 \frac{\log n}{n} + \sqrt{\frac{\log n}{n}} 8\beta.$$

As $v^* = \max_{\|\omega\|=1} \phi(\omega) \geq \phi(e_1)$, from Lemma 2.1 we obtain a lower bound of $v^*$ as presented below.

**Corollary 2.1** *For any $\delta > 0$, when $n \geq \max\{80, \frac{24}{\delta^2} \log \frac{1}{\delta}\}$, with probability at least $1 - 2n^{-8}$, $\sqrt{v^*} = \sqrt{\max_{\|\omega\|=1} \phi(\omega)}$ is lower bounded by $\beta - 4\delta$.*

The upper bound of $v^*$ is given in the follow lemma.

**Lemma 2.2** *There exist two numerical constants $C_4, C_5 \geq 4$ such that the following holds: for any $m \geq 1$, when $n \geq C_4 \frac{1}{\delta^2}(d \log \frac{1}{\delta} + m)$, with probability at least $1 - 3n^{-8} - \exp(-m)$,*

$$v^* = \max_{\|\omega\|=1} \phi(\omega) \leq (\beta + C_5 \delta)^2.$$

Let $\epsilon = C_5\delta$, then the assumptions of $n$ in Lemma 2.2 and Corollary 2.1 hold if

$$n \geq Cd\frac{1}{\epsilon^2}(\log\frac{1}{\epsilon} + m),$$

for some numerical constant $C$[2]. Combining Corollary 2.1 and Lemma 2.2, we obtain that with probability at least $1 - 5n^{-8} - \exp(-m)$,

$$|\sqrt{v^*} - \beta| \leq \epsilon,$$

which proves the claim. ■

**Remark:** Note that we allow $\epsilon > 1$, in which case $\log\frac{1}{\epsilon}$ is a negative number. In this case, the number of samples needed is independent of $\epsilon$ as shown in Corollary 2.1.

## 2.3. Proof of Lemma 2.1

To compute $\phi(e_1)$, we only need to consider the first components of $x^i, y^i$, *i.e.*, $x_1^i, y_1^i$, and ignore other components of them. Note that $\hat{x} = (x_1^1, \ldots, x_1^n)$ are i.i.d. samples from $\mathcal{N}(\beta, 1)$, and $\hat{y} = (y_1^1, \ldots, y_1^n)$ are i.i.d. samples from $\mathcal{N}(0, 1)$. Thus $\hat{x}_{[i]} - \beta, \hat{y}_{[i]}$ are order statistics of the standard Gaussian distribution. Let $Z_i, i \in \{1, \ldots, n\}$ be the order statistics of the standard Gaussian distribution $\mathcal{N}(0, 1)$. To simplify the notations, we let

$$u_i = \hat{x}_i - \beta,\ v_i = \hat{y}_i, 1 \leq i \leq n.$$

Consequently,

$$u_{[1]} \geq \cdots \geq u_{[n]} \quad \text{and} \quad v_{[1]} \geq \cdots \geq v_{[n]}$$

are ordered statistics of the standard Gaussian distribution, *i.e.*, realizations of $(Z_i)_{i=1}^n$. We have

$$\begin{aligned} n\phi(e_1) &= \sum_i (\hat{x}_{[i]} - \hat{y}_{[i]})^2 = \sum_i (u_{[i]} + \beta - v_{[i]})^2 \\ &= \sum_i (u_{[i]} - v_{[i]})^2 + 2\beta\sum_i (u_{[i]} - v_{[i]}) + n\beta^2. \end{aligned}$$

To get a sense about the magnitude this expression, we calculate the expectation as

$$\mathbb{E}(n\phi(e_1)) = \mathbb{E}\left(\sum_i (u_{[i]} - v_{[i]})^2\right) + 0 + n\beta^2 = 2(\sum_i \mathbb{E}(Z_i^2) - (\mathbb{E}\ Z_i)^2) + n\beta^2 = 2\sum_i \text{Var}(Z_i) + n\beta^2.$$

Obviously $\text{Var}(Z_k) = \text{Var}(Z_{n+1-k})$ due to symmetry of the standard Gaussian distribution, so we only need to consider $\text{Var}(Z_k)$ for $1 \leq k \leq n/2$. According to Proposition 4.2 of [1], the variance of $Z_k$ is bounded as $\text{Var}(Z_k) \leq C_0\frac{1}{k\log 2}$ for $1 \leq k \leq n/2$, where $C_0$ is a certain numerical constant. More specifically, note that the original presentation of Prop. 4.2 of [1] is for the absolute value of order statistics of Gaussian variables, denoted as $Y_{[k]}$, but we argue that the same proof applies to order statistics of the standard Gaussian distribution easily. Denote $Y_k$ as the absolute value of a standard Gaussian random variable. Note that the main part of the proof of Proposition 4.2 is based on Proposition 4.1, which estimates the hazard rate of $Y_k$. The same estimate also holds for the original Gaussian distribution; in fact, the proof of Proposition 4.1 (i) is essentially proving that the standard Gaussian distribution has non-decreasing hazard rate. Thus the whole Proposition 4.1 can be transformed to a version for the standard Gaussian distribution, except a possible constant factor in the bounds. The next step of the proof is to apply Theorem 2.9 which bounds the variance of $Z_k$ (recall that this is the order statistics of the standard Gaussian variable) by the hazard rate of the distribution; as the hazard rate estimate is given in Proposition 4.1, the same procedure in the proof of Prop. 4.2 leads to similar bounds for the variance of each $Z_k$ directly. Based on this estimate of the variance of $Z_k$, we have

$$\frac{1}{n}\sum_i \text{Var}(Z_i) \leq C_1\frac{\log n}{n},$$

where $C_1$ is a certain numerical constant. This implies that

$$\beta^2 \leq \mathbb{E}(\phi(e_1)) \leq 2C_1\frac{\log n}{n} + \beta^2.$$

[2]Note that the constant 80 in Corollary 2.1 and the gap between $\log\frac{1}{\delta} = \log\frac{1}{\delta} - \log C_5$ and $\log\frac{1}{\delta}$ can both be covered by a large enough numerical constant $C$.

We then prove that $\phi(e_1)$ concentrates around its expectation. We will apply McDiarmid's inequality for a Lipschitz function of Gaussian variables (see, *e.g.*, Theorem 2 of [3]). Define functions

$$F_1(u,v) = \sqrt{\sum_i (u_{[i]} - v_{[i]})^2} = \|u_{\text{ord}} - v_{\text{ord}}\|, \quad F_2(u,v) = 2\beta \sum_i (u_{[i]} - v_{[i]}) = 2\beta \sum_i (u_i - v_i).$$

where $u = (u_1, \dots, u_n)$, $v = (v_1, \dots, v_n)$ and $z_{\text{ord}}$ represents the reordered version of vector $z$. Then

$$\phi(e_1) = \frac{1}{n} F_1(u,v)^2 + \frac{1}{n} F_2(u,v) + \beta^2. \tag{2}$$

**Claim 2** *$\frac{1}{\sqrt{2}} F_1$ is 1-Lipschitz continuous with respect to the Euclidean metric.*

**Proof:** Consider another two vectors $a, b \in \mathbb{R}^n$. Then

$$F_1(u,v) - F_1(a,v) = \|u_{\text{ord}} - v_{\text{ord}}\| - \|a_{\text{ord}} - v_{\text{ord}}\| \le \|u_{\text{ord}} - a_{\text{ord}}\| \le \|u - a\|.$$

The last inequality holds because $\|u-a\|^2 - \|u_{\text{ord}} - a_{\text{ord}}\|^2 = 2\sum_i u_{[i]} a_{[i]} - 2\sum_i u_i a_i \ge 0$ due to the rearrangement inequality. Similarly,

$$F_1(a,v) - F_1(a,b) \le \|v - b\|.$$

Combining the above two inequalities, we have

$$F_1(u,v) - F_1(a,b) \le \|u-a\| + \|v-b\| \le \sqrt{2}\|(u;v) - (a;b)\|,$$

which proves this claim. □

According to McDiarmid's inequality for a Lipschitz function of Gaussian variables, we have

$$Pr(F_1(u,v)/\sqrt{2} - \mathbb{E}(F_1(u,v)/\sqrt{2}) \ge t) \le \exp(-t^2/\pi^2),$$

or

$$Pr(F_1(u,v) - \mathbb{E}(F_1(u,v)) \ge t) \le \exp(-\frac{t^2}{2\pi^2}).$$

As we already showed that $\mathbb{E}(F_1(u,v)^2) = 2\sum_i \text{Var}(Z_i) \le 2C_1 \log n$, we have

$$\mathbb{E}(F_1(u,v)) \le \sqrt{\mathbb{E}(F_1(u,v)^2)} \le \sqrt{2C_1}\sqrt{\log n}.$$

This implies

$$Pr(F_1(u,v) \ge t + \sqrt{2C_1}\sqrt{\log n}) \le \exp(-\frac{t^2}{2\pi^2}). \tag{3}$$

Letting $\epsilon_1 = \max\{\sqrt{2C_1}, 4\pi\}\sqrt{\log n}$, then

$$Pr(F_1(u,v) \ge 2\epsilon_1) \le Pr(F_1(u,v) - \mathbb{E}(F_1(u,v)) \ge \epsilon_1) \le \exp(-\frac{\epsilon_1^2}{2\pi^2}) \le \exp(\frac{-16\pi^2 \log n}{2\pi^2}) \le n^{-8}. \tag{4}$$

This implies that with probability at least $1 - n^{-8}$, we have

$$F_1(u,v)^2 \le 4\epsilon_1^2 = 4\max\{\sqrt{2C_1}, 4\pi\}^2 \log n = C_2 \log n, \tag{5}$$

where $C_2 = 4\max\{\sqrt{2C_1}, 4\pi\}^2$ is a numerical constant.

Bounding $F_2(u,v) = 2\beta\sum_i u_i - 2\beta\sum_i v_i$ is simple: it is just the sum of independent Gaussian variables. In fact, $\frac{1}{2\beta}F_2(u,v) = \sum_i u_i - \sum_i v_i \sim \mathcal{N}(0, 2n)$. By the standard Chernoff bound we have

$$Pr(|F_2(u,v)/(2\beta)| > t) \le \exp(-\frac{t^2}{2n}).$$

Let $t = \frac{8\sqrt{n \log n}}{2}$, then

$$Pr(|F_2| > 8\beta\sqrt{n \log n}) \leq \exp(-\frac{64n \log n}{8n}) = n^{-8}. \tag{6}$$

Combining Eq. (5) and Eq. (6), and the expression given in Eq. (2), with probability at least $1 - 2n^{-8}$, we have

$$\beta^2 - \frac{\log n}{\sqrt{n}} 8\beta \leq \phi(e_1) \leq \beta^2 + C_2 \frac{\log n}{n} + \sqrt{\frac{\log n}{n}} 8\beta, \tag{7}$$

which proves lemma 2.1. ■

**Remark**: The theoretical analysis implies that we need an increasing number of samples as the two distributions get close to each other. In particular, the number of samples required to estimate the distance is approximately $O(1/\beta^2)$, where $\beta$ is the distance between the two distributions.

## 2.4. Proof of Corollary 2.1

Assume this does not hold, $v^* < \beta - 4\delta$. Since $v^* \geq 0$, we have

$$\beta \geq 4\delta.$$

According to Lemma 2.1, we have

$$(\beta - 4\delta)^2 > (v^*)^2 \geq \beta^2 - 8\sqrt{\frac{\log n}{n}}\beta \qquad \Longrightarrow \qquad -8\sqrt{\frac{\log n}{n}}\beta < 16\delta^2 - 8\delta\beta \leq -4\delta\beta.$$

If $\beta = 0$, the above cannot hold, so $\beta > 0$. We then obtain $\delta < 2\sqrt{\frac{\log n}{n}}$, which implies

$$n < \frac{4}{\delta^2} \log n.$$

Define a function $\psi(x) = \frac{x}{\log x}$, then the above relation can be written as

$$\psi(n) < \frac{4}{\delta^2}. \tag{8}$$

Its derivative $\psi'(x) = \frac{\log x - 1}{\log(x)^2}$ which is positive for $x \geq 3$, thus $\psi(x)$ is strictly increasing in $[3, \infty)$.

If $\delta \geq 21/10$, then Eq. (8) implies $\psi(n) \leq 18$; since $\psi(80) > 18$, we have $n < 80$, which contradicts the assumption that $n > 80$. Therefore, we have

$$\delta < 21/10. \tag{9}$$

According to the assumption, we have

$$n \geq \frac{24}{\delta^2} \log \frac{1}{\delta}. \tag{10}$$

According to Eq. (8) and the monotonicity of $\psi$, we have

$$\begin{aligned} & \frac{4}{\delta^2} > \psi(n) \geq \psi(\frac{24}{\delta^2} \log \frac{1}{\delta}) \\ \Longrightarrow \quad & \frac{4}{\delta^2} \log(\frac{24}{\delta^2} \log \frac{1}{\delta}) \geq \frac{24}{\delta^2} \log \frac{1}{\delta} \\ \Longrightarrow \quad & \log(\frac{1}{\delta^2}) + \log(24 \log \frac{1}{\delta}) \geq 6 \log \frac{1}{\delta} \\ \Longrightarrow \quad & \log(24 \log \frac{1}{\delta}) \geq \log \frac{1}{\delta^4} \\ \Longrightarrow \quad & 24 \log \frac{1}{\delta} \geq \frac{1}{\delta^4} \\ \Longrightarrow \quad & \frac{21}{10} > \delta, \end{aligned}$$

which contradicts the assumption given Eq. (9) that $\delta \leq 21/10$. Consequently, corollary 2.1 holds. ■

## 2.5. Proof of Lemma 2.2

### 2.5.1 Preliminary Analysis

Recall that

$$v^* = \max_{\|\omega\|=1} \phi(\omega) = \max_{\|\omega\|=1} \frac{1}{n} \sum_i (\hat{x}_{[i]} - \hat{y}_{[i]})^2 = \max_{\|\omega\|=1} \frac{1}{n} \|(\omega^T X)_{\text{ord}} - (\omega^T Y)_{\text{ord}}\|^2,$$

where $z_{\text{ord}}$ represents the ordered version of vector $z$ such that the elements are non-decreasing and matrices $X = [x^1, \ldots, x^n]$, $Y = [y^1, \ldots, y^n]$. Moreover recall that $\hat{x} = (\omega^T x^1, \ldots, \omega^T x^n)$ are independent Gaussian random variables drawn from the distribution $\mathcal{N}(\beta\omega_1, 1)$, and $\hat{y} = (\omega^T y^1, \ldots, \omega^T y^n)$ are independent Gaussian random variables drawn from the distribution $\mathcal{N}(0,1)$, while $z_{[1]} \ge z_{[2]} \ge \cdots \ge z_{[n]}$ denotes the sorted elements of any vector $z$. Define $\tilde{\beta} = \omega_1\beta \le \beta$ (which is because $|\omega_1| \le \|\omega\| = 1$). According to Lemma 2.1, with probability at least $1 - 2n^{-8}$,

$$\phi(\omega) \le \tilde{\beta}^2 + C_2\frac{\log n}{n} + \sqrt{\frac{\log n}{n}} 8\tilde{\beta} \le \beta^2 + C_2\frac{\log n}{n} + \sqrt{\frac{\log n}{n}} 8\beta. \tag{11}$$

However, in the above argument, we have to first fix $\omega$ and then randomly sample $x^i, y^i$. The above argument does not show that with high probability Eq. (11) holds for *any* $\omega$. One standard method to resolve this issue is to use a covering argument, *i.e.*, first construct a $\delta$-covering $\{\omega^1, \ldots, \omega^N\}$ of the unit ball surface $\|\omega\| = 1$ so that each $\omega$ is close to some $\omega^j$, and then take the union bound to show that Eq. (11) holds for all $\omega^i$ with probability $1 - 2Nn^{-8}$, and finally bound the gap between the value at each point $\omega$ and the value at a close anchor point $\omega^j$. A formal proof is given below.

### 2.5.2 Formal Proof

We can decompose $\phi(\omega)$ as follows: let $\tilde{x}^i = x^i - \beta e_1$, $i = 1, \ldots, n$, and $\tilde{X} = (\tilde{x}^1, \ldots, \tilde{x}^n)$, then

$$\begin{aligned}
\phi(\omega) &= \frac{1}{n}\|(\omega^T X)_{\text{ord}} - (\omega^T Y)_{\text{ord}}\|^2 = \frac{1}{n}\|(\omega^T \tilde{X})_{\text{ord}} + \omega_1\beta - (\omega^T Y)_{\text{ord}}\|^2 \\
&= \frac{1}{n}\|(\omega^T \tilde{X})_{\text{ord}} - (\omega^T Y)_{\text{ord}}\|^2 + \frac{2}{n}\beta\omega_1[(\omega^T \tilde{X})_{\text{ord}} - (\omega^T Y)_{\text{ord}}] + \omega_1^2\beta^2 \\
&= \frac{1}{n}\|(\omega^T \tilde{X})_{\text{ord}} - (\omega^T Y)_{\text{ord}}\|^2 + \frac{2}{n}\beta\omega_1\omega^T(\tilde{X} - Y) + \omega_1^2\beta^2.
\end{aligned}$$

Define functions

$$F_1(\omega) = \|(\omega^T \tilde{X})_{\text{ord}} - (\omega^T Y)_{\text{ord}}\|, \quad F_2(\omega) = \omega^T(\tilde{X} - Y) = \omega^T \sum_i (\tilde{x}^i - y^i). \tag{12}$$

Then

$$\phi(\omega) = \frac{1}{n}F_1(\omega)^2 + \frac{2}{n}\beta\omega_1 F_2(\omega) + \omega_1^2\beta^2. \tag{13}$$

For fixed $\omega$, $\omega^T\tilde{x}^i, \omega^T\tilde{y}^i$ are independent standard Gaussian variables, thus $F_1(\omega) \le t + \sqrt{2C_1}\sqrt{\log n}$ with probability at least $1 - \exp(\frac{-t^2}{2\pi^2})$. To bound $F_1(\omega)$ for arbitrary $\omega$, we consider a $\delta$-covering of the $d$-dimensional unit ball sphere $\omega^1, \omega^2, \ldots, \omega^N$, which satisfies that for any $\|\omega\| = 1$, there exists some $\omega^k$ such that $\|\omega - \omega^k\| \le \delta$. It is not hard to argue that $N = O(1/\delta^d)$ points are enough. Then with probability at least $1 - N\exp(\frac{-t^2}{2\pi^2})$, we have $F_1(\omega^j) \le t + \sqrt{2C_1}\sqrt{\log n}$, $j = 1, \ldots, N$.

We then bound the gap between $F_1(\omega)$ and the value at some anchor point $F_1(\omega^j)$. For a certain $\omega^j$, suppose the ordering of the elements of $(\omega^j)^T\tilde{X}$ and $(\omega^j)^T Y$ are $\pi_1, \ldots, \pi_n$ and $\sigma_1, \ldots, \sigma_n$, *i.e.*,

$$\langle\omega^j, \tilde{x}_{\pi_1}\rangle \ge \cdots \ge \langle\omega^j, \tilde{x}_{\pi_n}\rangle; \quad \langle\omega^j, y_{\sigma_1}\rangle \ge \cdots \ge \langle\omega^j, y_{\sigma_n}\rangle.$$

Let $\tilde{X}_\pi = (\tilde{x}^{\pi_1}, \dots, \tilde{x}^{\pi_n}\})$ and $Y_\sigma = (y^{\sigma_1}, \dots, y^{\sigma_n})$, then we have

$$\begin{aligned}
F_1(\omega) - F_1(\omega^j) &= \|(\omega^T\tilde{X})_{\text{ord}} - (\omega^T Y)_{\text{ord}}\| - \|((\omega^j)^T\tilde{X})_{\text{ord}} - ((\omega^j)^T Y)_{\text{ord}}\| \\
&= \|(\omega^T\tilde{X})_{\text{ord}} - (\omega^T Y)_{\text{ord}}\| - \|(\omega^j)^T\tilde{X}_\pi - (\omega^j)^T Y_\sigma\| \\
&\leq \|\omega^T\tilde{X}_\pi - \omega^T Y_\sigma\| - \|(\omega^j)^T\tilde{X}_\pi - (\omega^j)^T Y_\sigma\| \\
&\leq \|\omega^T\tilde{X}_\pi - (\omega^j)^T\tilde{X}_\pi\| + \|\omega^T Y_\sigma - (\omega^j)^T Y_\sigma\| \\
&\leq \|\omega - \omega^j\|\sqrt{\lambda_{\max}(\tilde{X}_\pi\tilde{X}_\pi^T)} + \|\omega - \omega^j\|\sqrt{\lambda_{\max}(Y_\sigma Y_\sigma^T)} \\
&= \|\omega - \omega^j\|\left(\sqrt{\lambda_{\max}(\tilde{X}\tilde{X}^T)} + \sqrt{\lambda_{\max}(YY^T)}\right)
\end{aligned}$$

Note that $YY^T = \sum_{i,j} y^j(y^i)^T$ where each $y^i$ is a random Gaussian vector. Thus with probability at least $1-n^{-8}$, we have $\lambda_{\max}(YY^T) \leq C_3 n$ for some numerical constant $C_3$. Similarly, $\lambda_{\max}(\tilde{X}\tilde{X}^T) \leq C_3 n$ with probability at least $1-n^{-8}$. Thus with probability at least $1-2n^{-8}$, we have

$$F_1(\omega) - F_1(\omega^j) \leq \|\omega - \omega^j\| 2\sqrt{C_3 n}, \quad \forall \omega, \forall j \in \{1, \dots, N\}.$$

We then combine the previous two parts. For any $\omega$, there exists some $\omega^k$ such that $\|\omega - \omega^k\| \leq \delta$, thus

$$F_1(\omega) \leq F_1(\omega) - F_1(\omega^k) + F_1(\omega^k) \leq 2\sqrt{C_3 n}\delta + t + \sqrt{2C_1}\sqrt{\log n},$$

with probability at least $1 - N\exp(\frac{-t^2}{2\pi^2}) - 2n^{-8}$. Let $t = \sqrt{n}\delta$, then

$$F_1(\omega) \leq (2\sqrt{C_3} + 2)\sqrt{n}\delta + \sqrt{2C_1}\sqrt{\log n}, \tag{14}$$

with probability at least $1 - \frac{1}{\delta^d}\exp(\frac{-n\delta^2}{2\pi^2}) - 2n^{-8}$. Pick large enough $n$ such that

$$n \geq C_4 \frac{1}{\delta^2}(\log(1/\delta)d + m), \tag{15}$$

where $C_4 = \max\{2\pi^2, (\frac{\sqrt{C_1}}{\sqrt{2}(\sqrt{C_3}+1)})^2\}$, then it is easy to verify that (similar to the proof of Corollary 2.1) that

$$n \geq \max\left\{\frac{2\pi^2}{\delta^2}(\log(1/\delta)d + m),\ C_4\frac{1}{\delta^2}d\log n\right\}. \tag{16}$$

This relation further implies the following two relations:

$$\frac{1}{\delta^d}\exp(\frac{-n\delta^2}{2\pi^2}) = \exp(\frac{-n\delta^2}{2\pi^2} + d\log(1/\delta)) \leq \exp(-m),$$

and

$$\sqrt{\frac{\log n}{n}} \leq \sqrt{1/C_4}\delta \leq \frac{\sqrt{2}(\sqrt{C_3}+1)}{\sqrt{C_1}}\delta \quad \Longrightarrow \quad \sqrt{2C_1}\sqrt{\log n} \leq (2\sqrt{C_3}+2)\sqrt{n}.$$

Therefore, from Eq. (14), and by letting $C_5 = 4(\sqrt{C_3}+1)$, we obtain that

$$F_1(w) \leq C_5\sqrt{n}\delta \tag{17}$$

holds with probability at least $1 - \exp(-m) - 2n^{-8}$.

We then bound $F_2(\omega)$: by the fact that $\sum_i(\tilde{x}^i - y^i)$ is the sum of the 2nd standard Gaussian variables and by the standard Chernoff bound, with probability at least $1-n^{-8}$,

$$\max_{\|\omega\|=1} F_2(\omega) = \max_{\|\omega\|=1} \omega^T\sum_i(\tilde{x}^i - y^i) = \|\sum_i(\tilde{x}^i - y^i)\| \leq 4\sqrt{nd\log n}.$$

Combining the bounds of $F_1$ and $F_2$, and using the fact that $|\omega_1| \le 1$, we conclude that with probability at least $1 - \exp(-d) - 3n^{-8}$,

$$\max_{\|\omega\|=1} \phi(\omega) = \max_{\|\omega\|=1} \left( \frac{1}{n} F_1(\omega)^2 + \frac{2}{n}\beta\omega_1 F_2(\omega) + \omega_1^2\beta^2 \right) \le C_5^2\delta^2 + \frac{8}{n}\beta\sqrt{nd\log n} + \beta^2,$$

which implies

$$\max_{\|\omega\|=1} \sqrt{\phi(\omega)} \le \beta + \max\left\{ C_5\delta, 4\sqrt{\frac{d\log n}{n}} \right\}.$$

According to Eq. (16), we have

$$4\sqrt{\frac{d\log n}{n}} \le 4\frac{1}{\sqrt{C_4}}\delta \le \frac{4\sqrt{2}(\sqrt{C_3}+1)}{\sqrt{C_1}}\delta = \frac{2}{C_1}C_5\delta \le C_5\delta,$$

thus we can simplify the bound to

$$\max_{\|\omega\|=1} \sqrt{\phi(\omega)} \le \beta + C_5\delta,$$

which proves the lemma. ■

**Remark:** This lemma of the upper bound is loose, because we use the covering argument and the union bound. The reason for using the covering argument is because we want to upper bound $\phi(\omega)$ for all $\omega$. This proof strategy ignores the fact that there is only one $\omega^*$ (the optimal direction) that is effective, but introduces $N = 1/\delta^d$ anchor points. The outcome is that in Eq. (15) we need to introduce an extra multiplicative factor of $d$ and an extra additive factor of $m$. Stronger bounds are likely attainable.

### 2.6. Sliced Wasserstein distance is $\mathcal{P}$-generalizable

First, we give a formal statement of the desired result that the sliced Wasserstein distance is $\mathcal{P}$-generalizable. Recall that $\mu$ and $\nu$ are two distributions from the the family $\mathcal{P}$, and $\hat{\mu}, \hat{\nu}$ are empirical versions of $\mu, \nu$ each with size $n$.

**Proposition 2.2** *The sliced Wasserstein-2 distance is $\mathcal{P}$-generalizable in the following sense. Consider $K$ random directions $\omega^1, \ldots, \omega^K$ drawn from the Gaussian distribution $\mathcal{N}(0, I_d)$. Define the empirical distance as*

$$\tilde{W}_2(\hat{\mu}, \hat{\nu}) = \left[ \frac{1}{K} \sum_{k=1}^{K} W_2^2((\omega^k)^T\hat{\mu}, (\omega^k)^T\hat{\nu}) \right]^{\frac{1}{2}}.$$

*There exist some numerical constants $C_7, C_8, C_9$ such that if*

$$K \ge C_7\beta^2\frac{\log n}{\epsilon^2}, \quad n \ge C_8\frac{1}{\epsilon^2}\log\frac{1}{\epsilon},$$

*then with probability at least $1 - (5 + C_9\frac{\beta^2}{\epsilon^2})n^{-8}$, we have*

$$|\tilde{W}_2(\hat{\mu}, \hat{\nu}) - \tilde{W}_2(\mu, \nu)| < \epsilon.$$

**Remark:** The notion of generalization is somewhat different from the max-sliced Wasserstein distance: here we have requirements on both the number of projection directions and the number of samples.

**Proof:** Recall that we can assume $\nu \sim \mathcal{N}(0, I_d)$, $\mu \sim \mathcal{N}(\beta e_1, I_d)$ and $\beta \ge 0$. Also recall that we let $\hat{\mu}$ consist of vectors $x^1, \ldots, x^n$, and we let $\hat{\nu}$ consist of vectors $y^1, \ldots, y^n$. We define two matrices $X = [x^1, \ldots, x^n]$ and $Y = [y^1, \ldots, y^n]$.

For any unit vector $\omega \in \mathbb{R}^d$, we have $\omega^T\mu \sim \mathcal{N}(\beta\omega_1, 1)$. Similarly, for another Gaussian distribution $\nu \sim \mathcal{N}(0, I_d)$, we have $\omega^T\nu \sim \mathcal{N}(0, 1)$. The Wasserstein distance between the two distributions is

$$W_2(\omega^T\mu, \omega^T\nu) = \|\beta\omega_1 - 0\| = \beta|\omega_1|.$$

Therefore,

$$\tilde{W}_2(\mu, \nu) = \left[\mathbb{E}_{\omega\sim\mathcal{N}(0,I_d)} W_2^2(\omega^T\mu, \omega^T\nu)\right]^{\frac{1}{2}} = \left[\mathbb{E}_{\omega\sim\mathcal{N}(0,I_d)}\beta^2\omega_1^2\right]^{\frac{1}{2}} = \beta.$$

Recall that the empirical sliced Wasserstein distance is

$$\tilde{W}_2(\hat{\mu},\hat{\nu}) = \left[\frac{1}{K}\sum_{k=1}^{K} W_2^2((\omega^k)^T\hat{\mu},(\omega^k)^T\hat{\nu})\right]^{\frac{1}{2}}.$$

Let $v^* = \tilde{W}_2(\hat{\mu},\hat{\nu})^2$. For a fixed $\omega = e_1$, as shown in lemma 2.1, the following holds with probability at least $1 - n^{-8}$:

$$\beta^2 - \frac{\log n}{\sqrt{n}} 8\beta \le W_2^2(e_1^T\hat{\mu}, e_1^T\hat{\nu}) \le \beta^2 + C_2\frac{\log n}{n} + \sqrt{\frac{\log n}{n}} 8\beta. \tag{18}$$

For a fixed $\omega$, similarly, the following holds with probability at least $1 - n^{-8}$:

$$\omega_1^2\beta^2 - \frac{\log n}{\sqrt{n}} 8\beta\omega_1 \le W_2^2(\omega^T\hat{\mu}, \omega^T\hat{\nu}) \le \beta^2\omega_1^2 + C_2\frac{\log n}{n} + \sqrt{\frac{\log n}{n}} 8\beta\omega_1. \tag{19}$$

For $K$ fixed vectors $\omega^1,\dots,\omega^K$, by the union bound, with probability at least $1 - Kn^{-8}$,

$$\begin{aligned}\beta^2\frac{1}{K}\sum_{i=1}^{K}(\omega_1^i)^2 - \frac{\sqrt{\log n}}{\sqrt{n}} 8\beta\frac{1}{K}\sum_{i=1}^{K}\omega_1^i \le v^* \le \\ \beta^2\frac{1}{K}\sum_{i=1}^{K}(\omega_1^i)^2 + C_2\frac{\log n}{n} + \sqrt{\frac{\log n}{n}} 8\beta\frac{1}{K}\sum_{i=1}^{K}\omega_1^i.\end{aligned} \tag{20}$$

Note that the above holds for any fixed $\omega^1,\dots,\omega^K$. In our problem, these $\omega^1,\dots,\omega^K$ are random Gaussian vectors. We can view the process of determining the bounds of $\omega^k$ and $x^i, y^j$ as follows: we first randomly pick $\omega^1,\dots,\omega^K$ which will satisfy the bounds discussed below with high probability, then for these fixed directions we can apply the union bound and obtain the estimate in Eq. (20). This is valid as both $\omega^k$ and $x^i, y^j$ are independent.

Since $w_1^i, i \in \{1,\dots,n\}$ are standard Gaussian variables, we have with probability at least $1 - \exp(-t^2/2K)$,

$$1 - \frac{t}{K} \ge \frac{1}{K}\sum_{i=1}^{K}(\omega_1^i)^2 \le 1 + \frac{t}{K},$$

and with probability at least $1 - \exp(-t^2/K)$,

$$-\frac{t}{K} \ge \frac{1}{K}\sum_{i=1}^{K}\omega_1^i \le \frac{t}{K}.$$

We pick $t = 4\sqrt{K\log n}$, then with probability at least $1 - 2n^{-8}$,

$$1 - 4\frac{\sqrt{\log n}}{\sqrt{K}} \le \frac{1}{K}\sum_{i=1}^{K}(\omega_1^i)^2 \le 1 + 4\frac{\sqrt{\log n}}{\sqrt{K}}, \quad -4\frac{\sqrt{\log n}}{\sqrt{K}} \le \frac{1}{K}\sum_{i=1}^{K}\omega_1^i \le 4\frac{\sqrt{\log n}}{\sqrt{K}}.$$

Plugging into Eq. (20), we get

$$\begin{aligned}\beta^2(1 - 4\frac{\sqrt{\log n}}{\sqrt{K}}) - \frac{\sqrt{\log n}}{\sqrt{n}} 8\beta\frac{\sqrt{\log n}}{\sqrt{K}} \le v^* \le \\ \beta^2(1 + 4\frac{\sqrt{\log n}}{\sqrt{K}}) + C_2\frac{\log n}{n} + \sqrt{\frac{\log n}{n}} 8\beta\frac{\sqrt{\log n}}{\sqrt{K}}.\end{aligned} \tag{21}$$

Notice that there is a multiplicative error term $\frac{\sqrt{\log n}}{\sqrt{K}}\beta^2$. In the naïve case, in order to ensure $|\beta\sqrt{1 + 4\frac{\sqrt{\log n}}{\sqrt{K}}}) - \beta| < \epsilon$, we only need to pick $K \ge O(\frac{\beta^2\log n}{\epsilon^2})$. To handle the additive error term $\sqrt{\frac{\log n}{n} 8\beta\frac{\sqrt{\log n}}{\sqrt{K}}}$, the coefficient of $\beta$, which is $8\sqrt{\frac{\log n}{n}}\frac{\sqrt{\log n}}{\sqrt{K}}$, needs to be less than $\epsilon$, which holds under the condition $Kn \geq \frac{1}{\epsilon^2}\log^2 n$. If we pick $K \geq O(\log n)$ and $n \geq O(\frac{1}{\epsilon^2}\log\frac{1}{\epsilon})$, then the bound holds. Finally, the constant term error $C_2\frac{\log n}{n} < \epsilon^2$ requires $n > O(\frac{1}{\epsilon^2}\log\frac{1}{\epsilon})$.

In summary, to achieve error $\epsilon$, we need to pick some numerical constants $C_7, C_8, C_9$ and

$$K \geq C_7\beta^2\frac{\log n}{\epsilon^2}, n > C_8\frac{1}{\epsilon^2}\log\frac{1}{\epsilon},$$

then with probability at least $1-(5+C_9\frac{\beta^2}{\epsilon^2})n^{-8}$, we have

$$|\sqrt{v^*}-\beta| < \epsilon,$$

which proves the claim. ■

## 3. Proof of Claim 2 in the paper

In this part of the supplementary material we prove claim 2 of the paper.

**Claim 3** *(restated Claim 2 of the paper) The max-sliced Wasserstein-2 distance*

$$\textit{max-}\tilde{W}_2(\mu,\nu) = \left[\max_{\omega\in\Omega} W_2^2(\mu^\omega,\nu^\omega)\right]^{\frac{1}{2}}$$

*is a well defined distance between distributions.*

**Proof:** The conditions of non-negativity, symmetry, and identity of discernibles are trivially satisfied since the Wasserstein-2 distance is itself a well defined distance. The triangle inequality is proved as follows: Consider distributions $\mathbb{P}_1, \mathbb{P}_2$, and $\mathbb{P}_3$ over $\mathbb{R}^n$. We want to show that

$$\text{max-}\tilde{W}_2(\mathbb{P}_1,\mathbb{P}_2) \leq \text{max-}\tilde{W}_2(\mathbb{P}_1,\mathbb{P}_3) + \text{max-}\tilde{W}_2(\mathbb{P}_2,\mathbb{P}_3). \tag{22}$$

Suppose max-$\tilde{W}_2(\mathbb{P}_1,\mathbb{P}_2)$ is achieved along the projection direction $\omega^*$, *i.e.*,

$$\omega^* = \operatorname*{argmax}_{\omega\in\Omega} W_2^2(\mathbb{P}_1^\omega,\mathbb{P}_2^\omega). \tag{23}$$

Along $\omega^*$, by the triangle inequality of the Wasserstein-2 distance, we have

$$\text{max-}\tilde{W}_2(\mathbb{P}_1,\mathbb{P}_2) = W_2(\mathbb{P}_1^{\omega^*},\mathbb{P}_2^{\omega^*}) \leq W_2(\mathbb{P}_1^{\omega^*},\mathbb{P}_3^{\omega^*}) + W_2(\mathbb{P}_2^{\omega^*},\mathbb{P}_3^{\omega^*}). \tag{24}$$

From the definition of the max-sliced Wasserstein-2 distance and from Eq. (23) we obtain

$$W_2(\mathbb{P}_1^{\omega^*},\mathbb{P}_3^{\omega^*}) \leq \text{max-}\tilde{W}_2(\mathbb{P}_1,\mathbb{P}_3), \quad \text{and} \quad W_2(\mathbb{P}_2^{\omega^*},\mathbb{P}_3^{\omega^*}) \leq \text{max-}\tilde{W}_2(\mathbb{P}_2,\mathbb{P}_3). \tag{25}$$

Substituting Eq. (25) in Eq. (24) completes the proof. ■

## 4. Architecture for Image Generation

The architectures for experiments at an image resolution of 256x256 are described in Tab. 1 and Tab. 2. The architectures for an image resolution of 512x512 are described in Tab. 3 and Tab. 4. For all experiments, the Adam optimizer [2] with a learning rate of 0.0001 was used. The mini-batch size was set to 64 for 256x256 images, and 32 for images at 512x512.

## References

[1] S. Boucheron and M. Thomas. Concentration inequalities for order statistics. *Electronic Communications in Probability*, 2012. 5

[2] D. Kingma and J. Ba. Adam: A method for stochastic optimization. *arXiv preprint arXiv:1412.6980*, 2014. 12

[3] S. P. Lalley. Concentration inequalities. *Lecture notes, University of Chicago*, 2013. 6

[4] B. Laurent and P. Massart. Adaptive estimation of a quadratic functional by model selection. *Annals of Statistics*, pages 1302–1338, 2000. 2

| Layer | Stride | Output channels | Normalization | Activation |
|---|---|---|---|---|
| (Input) FC | N.A | (4x4x) 1024 | BN | Relu |
| Transpose Conv | 2 | 512 | BN | Relu |
| Transpose Conv | 2 | 256 | BN | Relu |
| Transpose Conv | 2 | 128 | BN | Relu |
| Transpose Conv | 2 | 64 | BN | Relu |
| Transpose Conv | 2 | 32 | BN | Relu |
| Transpose Conv | 2 | 16 | BN | Relu |
| Conv | 1 | 16 | BN | Relu |
| (Output) Conv | 1 | 3 | None | Tanh |

Table 1: Generator architecture for 256x256. BN = Batch normalization, LN = Layer normalization.

| Layer | Stride | Output channels | Normalization | Activation |
|---|---|---|---|---|
| Conv | 2 | 16 | LN | Relu |
| Conv | 2 | 32 | LN | Relu |
| Conv | 2 | 64 | LN | Relu |
| Conv | 2 | 128 | LN | Relu |
| Conv | 2 | 256 | LN | Relu |
| Conv | 2 | 512 | LN | Relu |
| (Output) FC | N.A | 1 | None | None |

Table 2: Discriminator architecture for 256x256. BN = Batch normalization, LN = Layer normalization.

| Layer | Stride | Output channels | Normalization | Activation |
|---|---|---|---|---|
| (Input) FC | N.A | (4x4x) 1024 | BN | Relu |
| Transpose Conv | 2 | 512 | BN | Relu |
| Transpose Conv | 2 | 256 | BN | Relu |
| Transpose Conv | 2 | 256 | BN | Relu |
| Transpose Conv | 2 | 128 | BN | Relu |
| Transpose Conv | 2 | 64 | BN | Relu |
| Transpose Conv | 2 | 32 | BN | Relu |
| Transpose Conv | 2 | 16 | BN | Relu |
| Conv | 1 | 16 | BN | Relu |
| Conv + Concat(previous) | 1 | 8 | BN | Relu |
| Conv + Concat(previous) | 1 | 4 | BN | Relu |
| (Output) Conv | 1 | 3 | None | Tanh |

Table 3: Generator architecture for 512x512. BN = Batch normalization, LN = Layer normalization.

| Layer | Stride | Output channels | Normalization | Activation |
|---|---|---|---|---|
| Conv | 2 | 16 | LN | Relu |
| Conv | 2 | 32 | LN | Relu |
| Conv | 2 | 64 | LN | Relu |
| Conv | 2 | 128 | LN | Relu |
| Conv | 2 | 128 | LN | Relu |
| Conv | 2 | 256 | LN | Relu |
| Conv | 2 | 512 | LN | Relu |
| (Output) FC | N.A | 1 | None | None |

Table 4: Discriminator architecture for 512x512. BN = Batch normalization, LN = Layer normalization.